\documentclass[conference]{IEEEtran}
\IEEEoverridecommandlockouts
\usepackage{cite}
\usepackage{amsmath,amssymb,amsfonts}
\usepackage{algorithmic}
\usepackage{graphicx}
\usepackage{threeparttable}
\usepackage{textcomp}
\usepackage{xcolor}
\usepackage{tabularx}
\colorlet{shadecolor}{gray!40}
\usepackage{tcolorbox}  

\makeatletter
\newcommand{\linebreakand}{%
  \end{@IEEEauthorhalign}
  \hfill\mbox{}\par
  \mbox{}\hfill\begin{@IEEEauthorhalign}
}
\makeatother

\definecolor{customDecreaseHigh}{HTML}{BAD19C}
\definecolor{customDecreaseMid}{HTML}{D2E3B7}
\definecolor{customDecreaseLow}{HTML}{E4F1CC}

\definecolor{customIncreaseHigh}{HTML}{F9AF8B}
\definecolor{customIncreaseMid}{HTML}{FBC8B8}
\definecolor{customIncreaseLow}{HTML}{FDD7A7}
\usepackage[T1]{fontenc}
\usepackage{graphicx}
\usepackage{booktabs}
\usepackage{multirow}
\usepackage{rotating}
\usepackage{amsmath}
\usepackage{float}

\usepackage[table]{xcolor}
\usepackage{threeparttable} 

\usepackage{hyperref}

\usepackage{bm}
\def\BibTeX{{\rm B\kern-.05em{\sc i\kern-.025em b}\kern-.08em
    T\kern-.1667em\lower.7ex\hbox{E}\kern-.125emX}}
\begin{document}

\title{Unveiling the Merits and Defects of LLMs in Automatic Review Generation for Scientific Papers
}

\author{
\IEEEauthorblockN{Ruochi Li}
\IEEEauthorblockA{\textit{North Carolina State University} \\
Raleigh, USA \\
rli14@ncsu.edu}
\and
\IEEEauthorblockN{Haoxuan Zhang}
\IEEEauthorblockA{\textit{University of North Texas} \\
Denton, USA \\
HaoxuanZhang@my.unt.edu}
\and
\IEEEauthorblockN{Edward Gehringer}
\IEEEauthorblockA{\textit{North Carolina State University} \\
Raleigh, USA \\
efg@ncsu.edu}
\linebreakand
\IEEEauthorblockN{Ting Xiao}
\IEEEauthorblockA{\textit{University of North Texas} \\
Denton, USA \\
Ting.Xiao@unt.edu}
\and
\IEEEauthorblockN{Junhua Ding}
\IEEEauthorblockA{\textit{University of North Texas} \\
Denton, USA \\
Junhua.Ding@unt.edu}
\and
\IEEEauthorblockN{Haihua Chen}
\IEEEauthorblockA{\textit{University of North Texas} \\
Denton, USA \\
Haihua.Chen@unt.edu}
}

\maketitle

\begin{abstract}
The surge in scientific submissions has placed increasing strain on the traditional peer-review process, prompting the exploration of large language models (LLMs) for automated review generation. While LLMs demonstrate competence in producing structured and coherent feedback, their capacity for critical reasoning, contextual grounding, and quality sensitivity remains limited. To systematically evaluate these aspects, we propose a comprehensive evaluation framework that integrates semantic similarity analysis and structured knowledge graph metrics to assess LLM-generated reviews against human-written counterparts. We construct a large-scale benchmark of 1,683 papers and 6,495 expert reviews from ICLR and NeurIPS in multiple years, and generate reviews using five LLMs. Our findings show that LLMs perform well in descriptive and affirmational content, capturing the main contributions and methodologies of the original work, with GPT-4o highlighted as an illustrative example, generating 15.74\% more entities than human reviewers in the strengths section of good papers in ICLR 2025. However, they consistently underperform in identifying weaknesses, raising substantive questions, and adjusting feedback based on paper quality. GPT-4o produces 59.42\% fewer entities than real reviewers in the weaknesses and increases node count by only 5.7\% from good to weak papers, compared to 50\% in human reviews. Similar trends are observed across all conferences, years, and models, providing empirical foundations for understanding the merits and defects of LLM-generated reviews and informing the development of future LLM-assisted reviewing tools. Data, code, and more detailed results are publicly available at \url{https://github.com/RichardLRC/Peer-Review}.
\end{abstract}


\section{Introduction}
The rapid advancement of artificial intelligence has catalyzed a surge in global research activity. This trend has led to an explosive growth in paper submissions to top-tier AI conferences, placing considerable strain on the existing peer review system~\cite{wang2020reviewrobot,wei2023academicgpt}. The efficiency and quality of the review process have been significantly compromised~\cite{thakkar2025llm}. For instance, ICLR conference experienced a 47\% increase in submission in 2024 and a further 61\% rise in 2025\footnote{The ICLR Submission Report: \url{https://tinyurl.com/2j9c54my}}, increasing the reviewing burden. To address this growing challenge, ICLR 2025\footnote{The ICLR Review Agents: \url{https://tinyurl.com/5269nvwd}} introduced an LLM-based feedback agent that provided optional suggestions to improve review clarity and usefulness, marking a practical step toward integrating LLMs into academic reviewing. Furthermore, AAAI conference\footnote{The AAAI Review Pilot Announcement: \url{https://tinyurl.com/3sz7t2jm}} has launched a pilot program for AAAI-26 that incorporates LLMs into the review process, using them to provide supplementary first-stage reviews and summarize reviewer discussions, while maintaining full human oversight to preserve scientific integrity. Meanwhile, LLM-assisted peer review automation has attracted increasing attention in the research community, with several studies developing frameworks based on prompt engineering, multi-agent architectures, and knowledge enhancement methods ~\cite{d2024marg,yu2024automated,Jin2024AgentReviewEP}. Despite demonstrating preliminary effectiveness, LLM-generated reviews exhibit substantial limitations under systematic evaluation. Recent assessments of content consistency, scoring accuracy, input robustness, and evaluation bias ~\cite{ning2024pico,Jin2024AgentReviewEP,shin2025mind,hossain2025llms} indicate that current models remain vulnerable to incomplete information and text manipulation. These limitations frequently manifest as seemingly plausible but empirically insufficient criticisms. Models are susceptible to biased inputs and often generate superficially reasonable feedback based on partial manuscript content ~\cite{ye2024we,zhou2024llm}.

Existing LLM evaluations focus predominantly on surface-level textual comparisons or behavioral imitation, largely neglecting deep semantic alignment between the generated review and the original paper. Although some prior work has examined knowledge structures within human-written reviews ~\cite{min2021predicting,wang2024content}, systematic comparative analyses between the knowledge completeness, contextual grounding, and critical reasoning abilities of LLM-generated and human-written reviews are still lacking. This analytical gap limits our understanding of whether LLMs truly ``comprehend'' the reviewed papers, thus hindering their integration into real-world scholarly communication.

To address this gap and build a more principled understanding of LLM behavior in scientific evaluation, we center our investigation on two critical questions: \textbf{(1) To what extent do LLMs understand the scientific papers they review?} \textbf{(2) What are the merits and defects of LLM-generated reviews compared to those written by human reviewers?}

The empirical findings from our analyses reveal several key behavioral patterns in LLM-generated reviews:
\begin{itemize}
    \item \textbf{Faithful description of affirmative content:}
    LLMs demonstrate strong reliability in summarizing affirmative aspects of a paper. In both the summary and the strengths sections, they consistently capture core contributions with high fidelity and frequently include more text content than human reviewers. Their responses exhibit close alignment with the original material, reflecting thorough surface-level comprehension.
    \item \textbf{Evaluative bias toward leniency:} LLMs tend to produce similar evaluations among papers of varying quality. Unlike human reviewers who adjust their feedback based on submission quality, LLMs are more likely to assign uniformly positive assessments, particularly overrating borderline and weak submissions.
    \item \textbf{Limited depth in critical analysis:}
    LLMs exhibit markedly lower conceptual richness in the evaluative components of the review, particularly in the weaknesses section. Compared to human reviewers, their outputs include fewer scientific entities, simpler relational structures, and less diversity in conceptual content, reflecting a more constrained capacity for analytical depth.
\end{itemize}

The contributions of this work are summarized as follows:
\begin{itemize}
    \item We construct a high-consensus benchmark dataset comprising 1,683 papers and 6,495 reviews from ICLR and NeurIPS in multiple years, with each paper assigned to a good, borderline, or weak category according to the consistency of reviewer agreement. To enable comparative evaluation, five SOTA LLMs were used to generate an equal number of reviews based on the same paper set.
    \item We propose a novel evaluation framework that combines semantic similarity analysis and structured knowledge graph methodologies, enabling comprehensive assessments of both surface-level comprehension and deeper conceptual analysis in LLM-generated reviews.
    \item Our evaluations reveal distinct behavioral patterns in LLM-generated reviews, highlighting reliable performance in summarizing affirmative content but significant limitations in critical analytical depth and evaluative discrimination.
\end{itemize}

\section{Related Work}

Recent advances in automated peer review have driven the creation of benchmark datasets and evaluation frameworks for assessing LLM-generated critiques. ReviewAdvisor ~\cite{yuan2022can} and SEA ~\cite{yu2024automated} respectively compiled large-scale datasets from venues including ICLR, NeurIPS, ACL, and CONLL, aimed at aspect-conditioned summarization and mismatch-based scoring. AgentReview ~\cite{Jin2024AgentReviewEP} simulated full review workflows with over 53,800 curated reviews, rebuttals, and meta-reviews from ICLR 2020–-2023. CycleResearcher ~\cite{weng2024cycleresearcher} introduced Review-5k with 4,991 papers and over 16,000 reviews focused on iterative review-revision dynamics. DeepReview ~\cite{zhu2025deepreview} annotated staged reasoning chains within 13,378 paper-review pairs, proposing the evidence-checked DeepReviewer 14B baseline model. Yu ~\cite{yu2025your} presented the largest corpus to date, covering 788,984 human and LLM-generated reviews from ICLR and NeurIPS from 2016 to 2024, using five major LLMs. While these datasets have advanced the field, they generally treat papers uniformly, without stratifying them by quality, which limits comparative insight. Our dataset introduces explicit categorization into good, borderline, and weak papers based on consistent human ratings, and includes reviews generated by five distinct LLMs for each paper. This enables systematic, quality-aware evaluation of model behavior across both submission standards and model architectures.

In addition to dataset construction, multiple studies have proposed evaluation frameworks for assessing LLM-generated reviews. Zhou ~\cite{zhou2024llm} and ReviewEval ~\cite{kirtani2025revieweval} utilized textual similarity metrics (e.g., ROUGE, BERTScore) combined with reasoning-based assessments through expert-written critiques or retrieval-based rebuttal simulations. Shin et al.~\cite{shin2025mind} introduced structured annotations for aspects like novelty and methodology, enabling topic-level alignment analysis. LLMetrica ~\cite{zhou2025large} integrated linguistic and semantic features with ScholarDetect classifiers, quantifying LLM involvement across reviews and abstracts. REMOR ~\cite{taechoyotin2025remor} employed reinforcement learning guided by surface-level metrics like METEOR to improve review-generation quality. While these methods provide valuable insights, they largely assess reviews at the whole-text level, lack section-specific evaluation, and ignore variations in paper quality. Our framework complements and extends these efforts through section-level semantic similarity analysis across quality tiers, enabling more precise and interpretable assessment.

Researchers have also explored knowledge-grounded methods for review generation and evaluation. Reviewer2 ~\cite{gao2024reviewer2} and KID-Review ~\cite{yuan2022kid} both emphasized aspect coverage and specificity, enriching LLM-generated critiques with paper-derived facets or knowledge graphs, and assessing aspect-level coverage and factual soundness. ReviewCritique ~\cite{du2024llms} proposed a human-annotated dataset specifically evaluating deficiencies and diversity within LLM-generated review content. Similarly, ReviewRobot ~\cite{wang2020reviewrobot} constructs Knowledge Graphs (KGs) from paper content and prior literature to generate evidence-grounded critiques. While these methods incorporate structured or conceptual representations, they either focus on generation or rely on coarse-grained aspect-level evaluation. In contrast, our framework uses automatic scientific entity and relation extraction to build paper and review-specific KGs, and employs interpretable graph metrics to assess conceptual alignment and granularity across different paper quality levels and model outputs.

\section{Methodology}


\begin{figure*}[t]
    \centering
    \includegraphics[width=\textwidth]{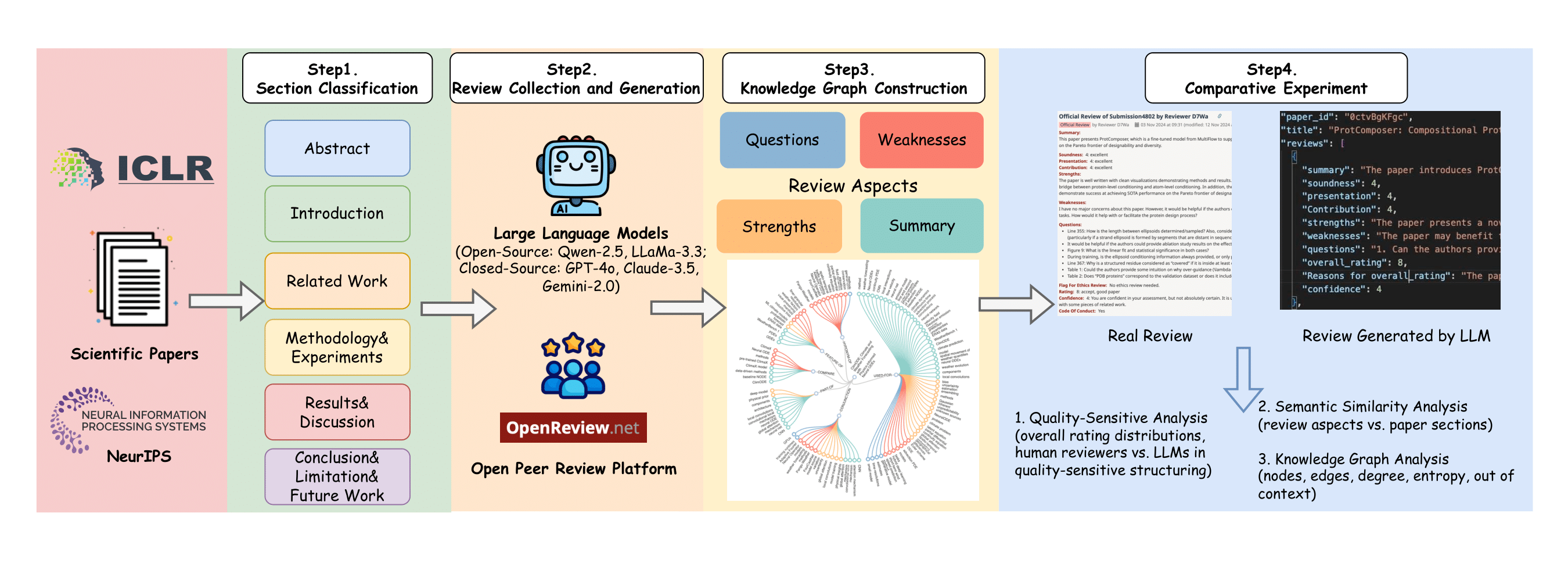} 
    \caption{The comprehensive framework for evaluating LLM-generated vs. human peer reviews: Benchmark construction from ICLR and NeurIPS papers, multi-model review generation with conference-specific criteria, and multi-dimensional comparative analysis using semantic similarity and knowledge graphs.}
    \label{fig:framework}
\end{figure*}

Figure~\ref{fig:framework} illustrates the overall framework of this research. We first construct a benchmark dataset that leverages two top-tier conferences (ICLR and NeurIPS) with varying paper quality levels (good, borderline, and weak). This dataset includes full-text papers and their corresponding human-written reviews from OpenReview, which offers structured and authentic review interactions continuously evolving across conference cycles~\cite{sun2025openreview}. Next, we use prompts based on different conference scoring criteria to generate reviews from various LLMs. In our comparative experiments, we apply quality-sensitive, semantic similarity analysis and knowledge graphs to examine differences. We compare authentic reviews against generated ones based on paper section structure and review aspects. This approach reveals both merits and defects of LLM-automated paper reviews in terms of similarity, structure, and knowledge content.

\subsection{Dataset Preparation} \label{Dataset Preparation}
To compare reviews written by real reviewers and those generated by LLMs, we construct a benchmark dataset based on publicly available peer reviews from ICLR and NeurIPS. These venues host the majority of submissions and official reviews on the OpenReview platform. We curated papers from ICLR 2024, 2025 and NeurIPS 2023, 2024, and applied a two-stage filtering process to ensure both reviewer agreement and diversity in paper quality.


\begin{figure}[ht]
    \centering
    \includegraphics[width=\linewidth]{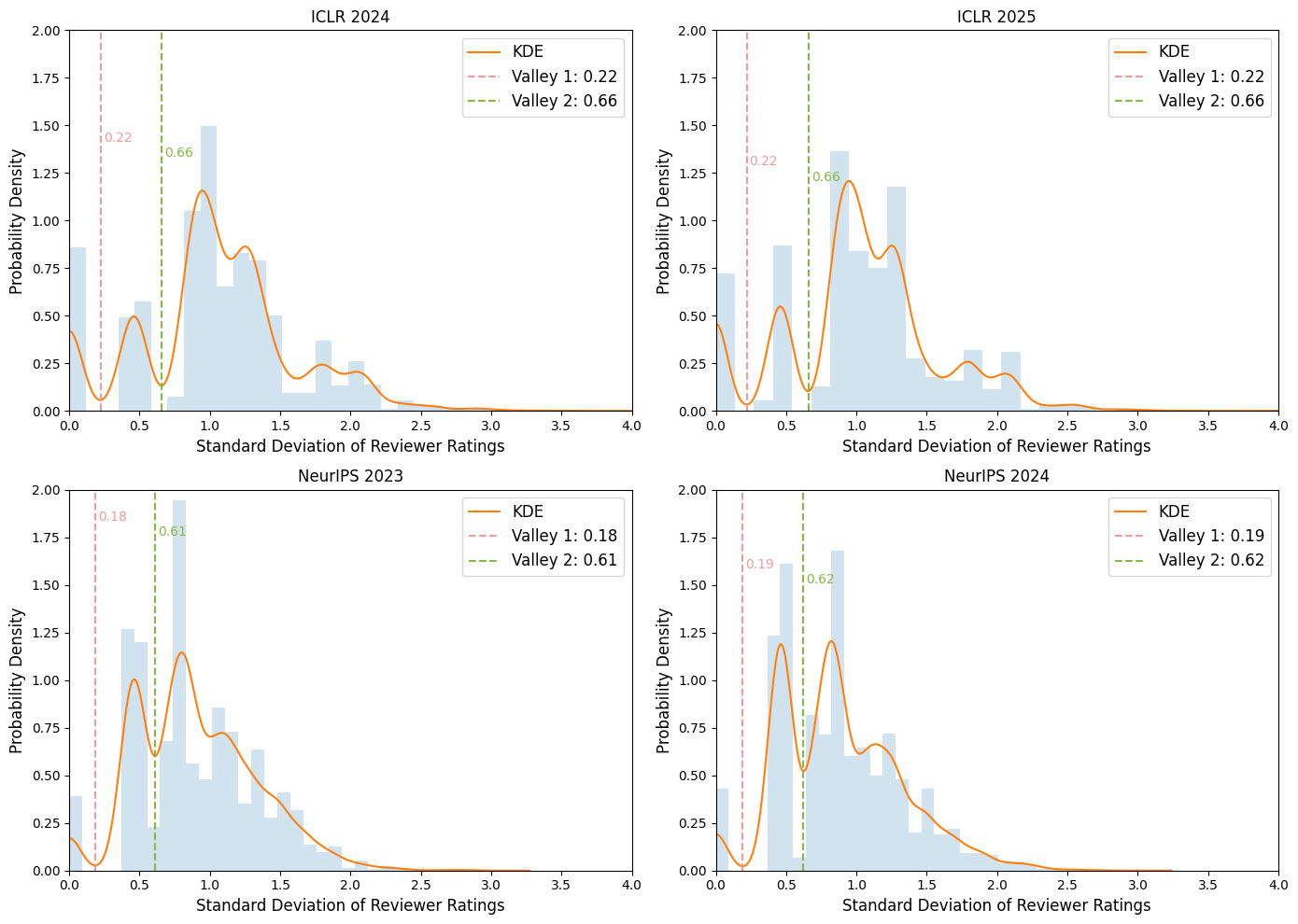}
    \caption{Kernel density estimation of review score standard deviations. The second local minimum (Valley 2) is chosen as the consistency threshold, capturing papers with strong but non-unanimous reviewer agreement to balance reliability and diversity.}
    \label{fig:Paper-Select}
\end{figure}

In the first stage, we aimed to identify papers with high reviewer agreement by selecting a consistency threshold. Prior work typically emphasizes the use of low-variance subsets to ensure reliable evaluations, but fails to specify how the threshold is determined ~\cite{xu2024good}. In contrast, we adopted a data-driven approach by leveraging \textit{kernel density estimation (KDE)}~\cite{silverman2018density}, a non-parametric method for estimating the probability density function of a variable, to model the empirical distribution of review score standard deviations. We applied KDE to compute a smooth density curve from all paper review scores' standard deviations and identified local minima on this curve to guide threshold selection. This allows us to retain informative variation while excluding noisy or highly inconsistent cases. We selected the second local minimum in the KDE curve as our consistency threshold. The first minimum, typically located at zero standard deviation, corresponds to absolute consistency among reviewers. However, we believe that peer reviews should allow for a certain degree of reasonable bias, as complete agreement may overlook diverse expert perspectives. Therefore, we chose the second minimum, which better reflects papers that exhibit strong yet non-unanimous reviewer agreement, enabling a balance between consistency and diversity. The threshold selection can be visualized in Figure~\ref{fig:Paper-Select}. Papers not meeting this threshold were excluded, as a large variance in reviewer scores reflects a lack of consensus among reviewers, making such papers unsuitable for reliable comparison with LLM-generated reviews.

\begin{table}[ht]
    \centering
    \caption{Number of Selected Papers with Review Counts by Conference, Year, and Quality Category}
    \renewcommand{\arraystretch}{0.5}
    \resizebox{0.5\textwidth}{!}{ \small
    \begin{tabular}{lccccccccc}
        \toprule
        \multirow{2}{*}{\textbf{Conference}} 
        & \multicolumn{3}{c}{\textbf{2023}} 
        & \multicolumn{3}{c}{\textbf{2024}} 
        & \multicolumn{3}{c}{\textbf{2025}} \\
        \cmidrule(lr){2-4} \cmidrule(lr){5-7} \cmidrule(lr){8-10}
        & Good & Border & weak 
        & Good & Border & weak 
        & Good & Border & weak \\
        \midrule
        ICLR     & - & - & -  & 84 & 200 & 350  & 98 & 308 & 301 \\
        NeurIPS  & 50 & 74 & 24  & 64 & 103 & 27  & - & - & - \\
        \midrule
        Reviews & 206 & 313 & 100  & 545 & 1749 & 812  & 369 & 1263 & 1138 \\
        \bottomrule
    \end{tabular}
    }
    \label{tab:paper-distribution}
\end{table}

In the second stage, we categorized papers into high-quality (good), mid-quality (borderline), and low-quality (weak) groups to ensure diversity in paper quality. We ranked papers by their aggregated review scores and selected the top 2.5\%, bottom 2.5\%, and middle 2.5\%, approximating $\mu \pm 2\sigma$ in a normal distribution to capture significant quality variations. The final dataset balances consistent, diverse reviews for comparing human and LLM-generated reviews. The rating distributions and category boundaries are shown in Figure~\ref{fig:overall_rating}, which provides further clarity on the score segmentation. The final dataset comprised 1,683 papers and 6,495 reviews, including 296 good papers, 685 borderline papers, and 702 weak papers. The detailed distribution is presented in Table~\ref{tab:paper-distribution}.

\subsection{Review Collection and Generation} \label{Review Collection and Generation}
To obtain authentic peer reviews, we systematically collected reviewer feedback for each selected paper from the OpenReview API. For each paper, we retrieved all reviews provided by assigned reviewers, encompassing both textual and numerical components. The textual feedback consists of the reviewer-written \textit{summary}, \textit{strengths}, \textit{weaknesses}, and \textit{questions}, while the numerical scores include evaluations of \textit{soundness}, \textit{presentation}, \textit{contribution}, \textit{overall rating}, and \textit{reviewer confidence}.

To ensure fairness and consistency in our comparative analysis, we generated LLM-based reviews using prompts shown in the following color box, which were explicitly derived from the official review rubrics of the ICLR and NeurIPS conferences\footnote{The ICLR Reviewer Guide: \url{https://tinyurl.com/bdeat2pu}.} \footnote{The NeurIPS Reviewer Guide: \url{https://tinyurl.com/5bw9p2e9}}. To standardize the input format for LLM processing, each paper was first converted from PDF to structured markdown using Nougat ~\cite{blecher2023nougat}, a Visual Transformer–based OCR system designed for scientific documents. These markdown-formatted papers were then concatenated with rubric-based prompts to provide clear and structured instructions to the LLMs. To ensure alignment with real reviews, each LLM was prompted to generate content matching the structure and quantity of the human-written reviews for each paper. 

{\small
\begin{tcolorbox}[colback=gray!5, colframe=black, boxrule=1pt, rounded corners, title=\textbf{Prompt template to generate reviews using LLMs based on fulltext and review guidelines}, fonttitle=\bfseries]

\textbf{System Prompt}\\ 
You are a professional academic paper reviewer. Evaluate papers based on the grading rubric provided. Return your response in JSON format with the following structure:\\
\vspace{2pt}
\begin{tabularx}{\linewidth}{@{}lX@{}}
\textit{Title:} & \textless \textit{Title of the paper}\textgreater \\
\textit{Summary:} & \textless \textit{Briefly summarize the paper and its contributions}\textgreater \\
\textit{Soundness:} & \textless \textit{Numerical rating from 1 to 4}\textgreater \\
\textit{Presentation:} & \textless \textit{Numerical rating from 1 to 4}\textgreater \\
\textit{Contribution:} & \textless \textit{Numerical rating from 1 to 4}\textgreater \\
\textit{Strengths:} & \textless \textit{Reasons you might accept the paper}\textgreater \\
\textit{Weaknesses:} & \textless \textit{Reasons you might reject the paper}\textgreater \\
\textit{Questions:} & \textless \textit{Questions and suggestions for the authors}\textgreater \\
\textit{Overall Rating:} & \textless \textit{Overall rating from 1 to 10}\textgreater \\
\textit{Reason for Rating:} & \textless \textit{Justification for the overall rating}\textgreater \\
\textit{Confidence:} & \textless \textit{Confidence rating from 1 to 5}\textgreater \\
\end{tabularx}

\vspace{4pt}
\textbf{User Prompt} \\
Here is the submitted manuscript: \textit{<Full-text Paper>}

\end{tcolorbox}
}

We selected five advanced LLMs for evaluation: \textbf{GPT-4o}, \textbf{Gemini-2.0-Flash}, \textbf{Claude-3.5-Sonnet}, \textbf{Qwen2.5-72B-instruct}, and \textbf{LLaMA3.3-70B-instruct}. The selection spans both proprietary and open-source models, reflecting different development paradigms, robust long-context processing capabilities essential for full-text comprehension, and SOTA performance on recent LLMs benchmarks.

\subsection{Comparative Metrics Framework Construction}\label{Comparative Metrics}

To systematically examine differences across multiple dimensions, we developed a comparative metrics framework that integrates semantic similarity analysis and knowledge graph methodologies. This framework enables us to compare (1) various sections within academic papers, (2) distinct aspects of peer review reports, (3) authentic human-written reviews versus those generated by LLMs, and (4) evaluations by different LLM architectures. By employing this comprehensive comparative approach, we elucidated both the merits and defects of LLM-based paper reviews, assessed from the dual perspectives of semantic similarity and knowledge representation.

\subsubsection{Semantic Similarity Metrics}\label{Semantic Similarity Metrics}

To assess the contextual relevance of peer reviews, we performed semantic similarity analysis between each section of the paper and each aspect of the review. This enables a quantitative assessment of how well the content of a review, whether written by a human or generated by an LLM, aligns with specific parts of the paper. To support this alignment, we segmented each paper into six sections following the IMRaD structure~\cite{sollaci2004introduction}: \textit{Abstract}, \textit{Introduction}, \textit{Related Work}, \textit{Methodology and Experiments}, \textit{Results and Discussions}, and \textit{Conclusion and Future Work}. This segmentation was automatically performed using the Qwen2.5-72B model on the markdown-formatted paper texts.

Each review contains four components: \textit{Summary}, \textit{Strengths}, \textit{Weaknesses}, and \textit{Questions}. Let \( R_i \) denote the \( i \)-th review component and \( S_j \) denote the \( j \)-th section of the paper. Using the BGE-M3 embedding model~\cite{chen2024bge}, we encoded both \( R_i \) and \( S_j \) into dense vector representations, and computed their semantic similarity via cosine similarity:
\begin{equation}
\text{sim}(R_i, S_j) = \frac{\mathbf{r}_i \cdot \mathbf{s}_j}{\|\mathbf{r}_i\| \|\mathbf{s}_j\|}
\end{equation}
This analysis allows direct comparison of semantic alignment across human and model-generated reviews, providing a fine-grained understanding of how each type of review reflects the actual content of the paper.

\subsubsection{Knowledge Graph Metrics}\label{Knowledge Graph Metrics}
In addition to semantic similarity analysis, we constructed KGs for each review to represent the scientific concepts and the semantic relationships expressed in the text. This structured representation complements surface-level similarity by capturing the conceptual scope and organization of the review content.

To construct the KG, we defined both entities and relations using the schema specified in the SciERC dataset ~\cite{luan2018multi}. Entity types included \textit{Task}, \textit{Method}, \textit{Metric}, \textit{Material}, \textit{Generic}, and \textit{Other Scientific Terms}. The relation types were drawn from \textit{part of}, \textit{used for}, \textit{feature of}, \textit{evaluate for}, \textit{hyponym of}, \textit{conjunction}, and \textit{compare}. This schema is well aligned with our research, as it captures a broad range of scientific concept types and relation structures commonly found in computer science literature. Its established role in previous work on scientific information extraction further supports its suitability for the construction of KGs ~\cite{min2021predicting,wang2024content}. To extract both entities and relations from the review text, we adopted the PL-Marker model ~\cite{ye2021packed}, which achieves SOTA performance on the SciERC benchmark and is trained to perform entity recognition along with relation extraction, making it a reliable backbone for KG construction.

Based on the resulting entity graph \( G = (V, E) \), where \( V \) denotes the set of extracted knowledge entities and \( E \) denotes the set of semantic relations between them, we computed a set of structural metrics to quantify the complexity and organization of the knowledge structure of each review:
\begin{itemize}
    \item {Number of nodes} \( |V| \): the total number of entities identified in the review section, reflecting the breadth of scientific content mentioned.

    \item {Number of edges} \( |E| \): the number of directed relations between entity pairs, representing how frequently the identified concepts are semantically connected.

    \item {Average degree}: the average number of relations per entity, indicating the density and interconnectivity of the graph. It is computed as:
    \begin{equation}
    \text{AvgDeg}(G) = \frac{1}{|V|} \sum_{v \in V} \deg(v)
    \end{equation}
    where \( \deg(v) \) denotes the total degree of node \( v \), defined as the sum of its in-degree and out-degree. The in-degree counts the number of edges pointing to \( v \), and the out-degree counts the number of edges originating from \( v \).

    \item {Label entropy} \( H(\mathcal{L}) \): the entropy of the entity type distribution, reflecting the diversity of entity categories present in the graph. Let \( \mathcal{C} \) be the set of entity categories and \( p(c) \) the proportion of entities labeled as type \( c \), then:
    \begin{equation}
    H(\mathcal{L}) = - \sum_{c \in \mathcal{C}} p(c) \log p(c)
    \end{equation}
    A higher entropy value indicates a more balanced and diverse representation of entity types, suggesting a broader conceptual scope.
\end{itemize}

We further assessed contextual grounding by examining whether each extracted entity appears in the original paper. Entities were grouped into in-context (present in the original paper) and out-of-context (absent from the original paper) categories. The relative sizes of these two sets reflect the extent to which the review content is supported by the source material. A higher proportion of in-context entities suggests strong textual fidelity, while more out-of-context entities may indicate the incorporation of external knowledge or inferred content.

This KG-based analysis provides a complementary perspective on review quality by quantifying both the conceptual coverage and the degree of source alignment.

\section{Experiment and Analysis}

\subsection{Experimental Setup}
For deployment, proprietary models (GPT-4o, Gemini-2.0-Flash, Claude-3.5-Sonnet) were accessed through official APIs with default generation parameters, while open-source models (Qwen2.5-72B-instruct, LLaMA3.3-70B-instruct) were locally served using vLLM with bf16 precision on two NVIDIA H100 NVL GPUs, each with 94 GB of memory. To accommodate memory constraints, we included the full main text of each paper and truncated only the appendix when necessary, using a token limit of 55k for Qwen2.5 and 65k for LLaMA3.3. This setup guarantees comprehensive coverage of the primary content from both ICLR and NeurIPS submissions.

\subsection{Quality-Sensitive Analysis}

To assess how well LLMs and human reviewers differentiate between papers of varying quality, we analyzed two complementary aspects: the distribution of overall ratings and the structural richness of the review content. While human reviews demonstrate clear quality awareness in both score assignment and content structure, LLM-generated reviews exhibit limited sensitivity to paper quality, especially for borderline and low-quality submissions.

\begin{itemize}
    \item \textbf{Overall rating distributions.} As shown in Figure~\ref{fig:overall_rating}, real reviewers produce well-separated rating distributions across good, borderline, and weak papers, indicating consistent calibration with paper quality. In contrast, LLMs tend to assign compressed and inflated scores, particularly overestimating borderline and weak submissions. This lack of separation suggests that current models have difficulty making nuanced score judgments.

    \begin{figure}[ht]
        \centering
        \includegraphics[width=\linewidth]{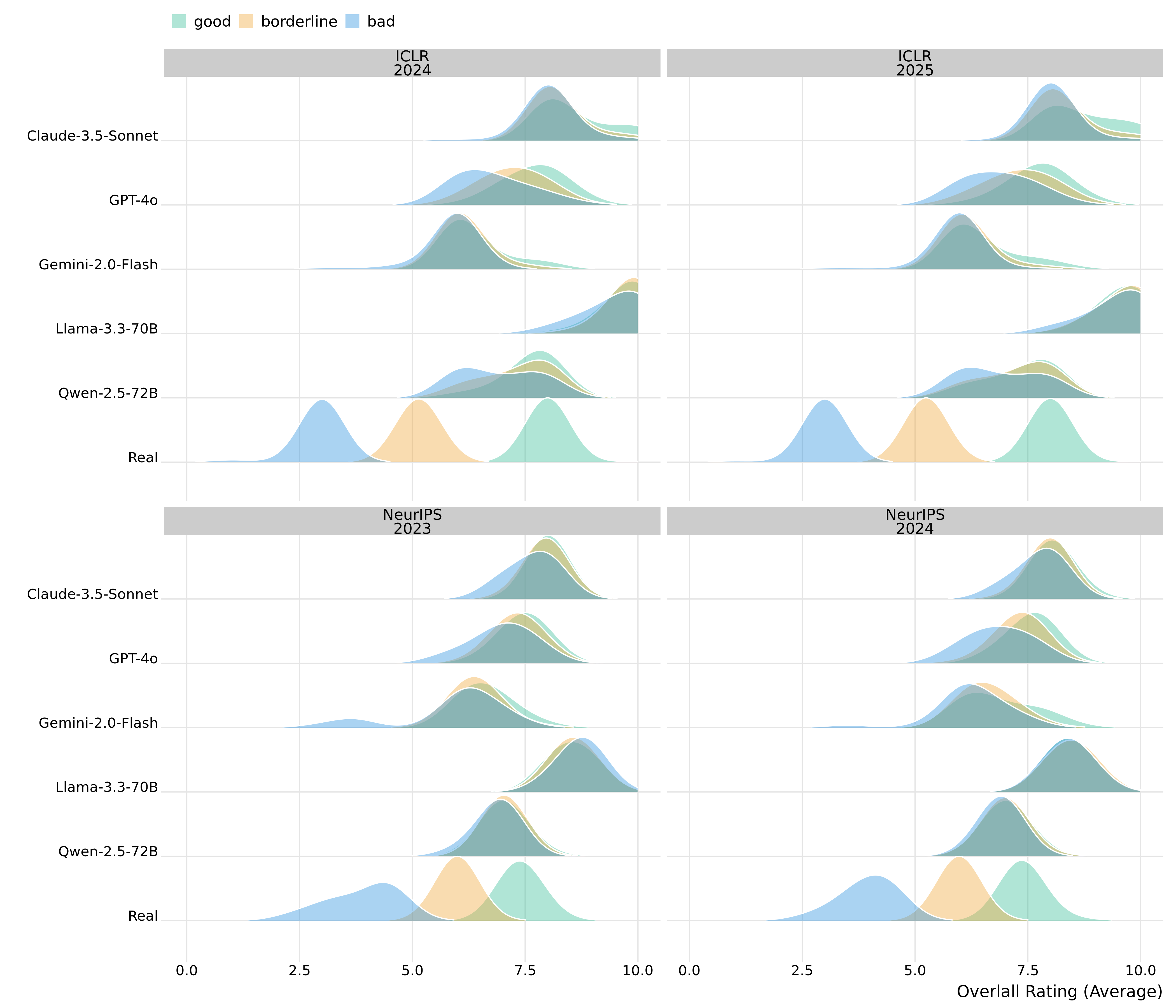}
        \caption{Overall rating distributions across paper quality categories}
        \label{fig:overall_rating}
    \end{figure}

    \item \textbf{Human reviewers vs. LLMs in quality-sensitive structuring.} Human-written reviews show clear trends in how structural content varies with submission quality. In the \textit{Weaknesses} section, the number of extracted entities increases with declining paper quality, suggesting that reviewers provide more detailed and diverse critical feedback for lower-quality submissions. For example, in the ICLR 2025 dataset in Table~\ref{ICLR and NIPS Table}, the number of nodes rises from 9.12 for good papers to 11.20 for borderline and 13.68 for weak papers, resulting in a 50.0\% increase. In contrast, the \textit{Strengths} section shows a decreasing trend, with node counts dropping from 6.04 to 4.46 and then to 3.47, a 42.5\% reduction, reflecting a reduction in affirmational content for weaker papers. This bidirectional gradient is consistently observed across the two conferences and years analyzed, indicating a quality-sensitive modulation in human-written reviews. By comparison, LLM-generated reviews exhibit minimal structural variation across paper quality levels, revealing a tendency to produce uniform outputs regardless of submission quality. For example, in the ICLR 2025 dataset, GPT-4o generates 3.70, 3.87, and 3.91 nodes in the \textit{Weaknesses} section for good, borderline, and weak papers, reflecting only a 5.7\% increase from good to weak. A similar flattening appears in the \textit{Strengths} section: while real reviews reduce affirmational content for lower-quality submissions, GPT-4o generates 6.99 for good, 7.29 for borderline, and 7.28 for weak papers, showing minimal sensitivity to paper quality.

    \begin{figure}[ht]
        \centering
        \includegraphics[width=\linewidth]{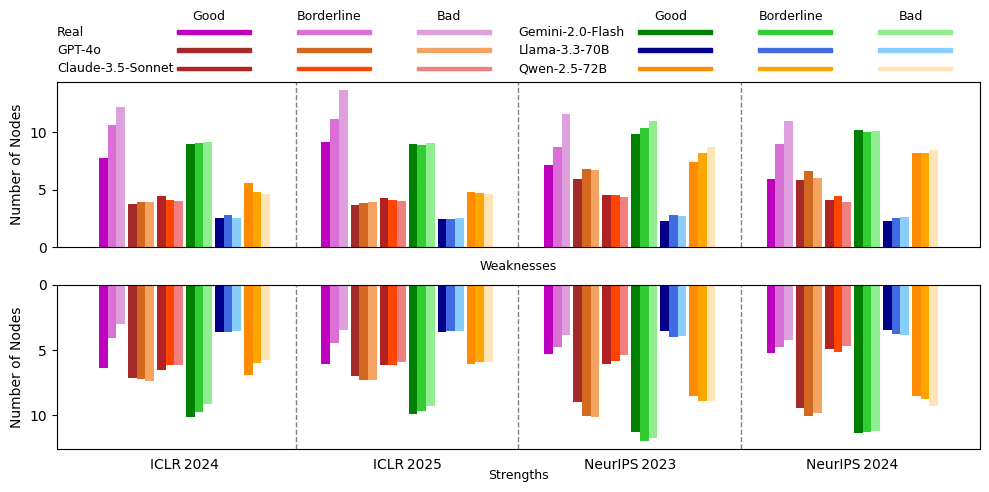}
        \caption{Number of extracted nodes in \textit{Weaknesses} and \textit{Strengths} across models and paper quality levels}
        \label{fig:num-of-node}
    \end{figure}
\end{itemize}

\begin{table*}[!ht]
\centering
\scriptsize
\renewcommand{\arraystretch}{1.2}
\setlength{\tabcolsep}{4pt}
\caption{Knowledge Graph Comparison for ICLR (2024, 2025) and NeurIPS (2023, 2024) Across LLMs and Sections (Relative Ratio to Real)}
\label{ICLR and NIPS Table}

\newcommand{\coloredcell}[1]{%
  \ifdim#1pt>75pt
    \cellcolor{customIncreaseHigh}#1\%%
  \else\ifdim#1pt>50pt
    \cellcolor{customIncreaseMid}#1\%%
  \else\ifdim#1pt>20pt
    \cellcolor{customIncreaseLow}#1\%%
  \else\ifdim#1pt<-75pt
    \cellcolor{customDecreaseHigh}#1\%%
  \else\ifdim#1pt<-50pt
    \cellcolor{customDecreaseMid}#1\%%
  \else\ifdim#1pt<-20pt
    \cellcolor{customDecreaseLow}#1\%%
  \else
    #1\%%
  \fi\fi\fi\fi\fi\fi
}

    \resizebox{\textwidth}{!}{
        \begin{tabular}{l | l |
            *{3}{c}| *{3}{c}| *{3}{c}| *{3}{c}| *{3}{c}| *{3}{c}| *{3}{c}
            }
            \toprule
            \textbf{Section} & \textbf{LLM} 
            & \multicolumn{3}{c|}{\textbf{Num Nodes}} 
            & \multicolumn{3}{c|}{\textbf{Num Edges}} 
            & \multicolumn{3}{c|}{\textbf{Avg Degree}} 
            & \multicolumn{3}{c}{\textbf{Label Entropy}}
            & \multicolumn{3}{c}{\textbf{In-Context Entities}}
            & \multicolumn{3}{c}{\textbf{Out-of-Context Entities}}
            & \multicolumn{3}{c}{\textbf{In-to-Out Ratio}}\\
            \midrule
            & & Good & Border & weak 
            & Good & Border & weak 
            & Good & Border & weak
            & Good & Border & weak 
            & Good & Border & weak
            & Good & Border & weak 
            & Good & Border & weak \\
            \midrule
            \multirow{6}{*}{\shortstack{Questions\\(ICLR 2024)}}
                & GPT-4o  & \coloredcell{-6.33} & \coloredcell{-5.31} & \coloredcell{-8.37} & \coloredcell{+24.13} & \coloredcell{+16.18} & \coloredcell{+25.17} & \coloredcell{+61.98} & \coloredcell{+56.04} & \coloredcell{+69.42} & \coloredcell{+34.89} & \coloredcell{+43.82} & \coloredcell{+42.84} & \coloredcell{+9.66} & \coloredcell{+25.99} & \coloredcell{+23.94} & \coloredcell{+5.74} & \coloredcell{+26.37} & \coloredcell{+18.58} & 66.09\% & 61.68\% & 60.10\% \\
                & Claude-3.5-Sonnet  & \coloredcell{-11.73} & \coloredcell{-8.68} & \coloredcell{-10.40} & \coloredcell{-19.02} & \coloredcell{-19.05} & \coloredcell{-17.08} & \coloredcell{+9.43} & \coloredcell{+13.74} & \coloredcell{+12.46} & \coloredcell{+37.77} & \coloredcell{+44.27} & \coloredcell{+49.70} & \coloredcell{+5.39} & \coloredcell{+23.18} & \coloredcell{+21.17} & \coloredcell{-3.61} & \coloredcell{+17.29} & \coloredcell{+16.68} & 62.69\% & 58.55\% & 60.48\% \\
                & Gemini-2.0-Flash  & \coloredcell{+136.60} & \coloredcell{+139.21} & \coloredcell{+128.84} & \coloredcell{+140.90} & \coloredcell{+139.14} & \coloredcell{+150.46} & \coloredcell{+36.14} & \coloredcell{+36.93} & \coloredcell{+47.23} & \coloredcell{+75.57} & \coloredcell{+79.32} & \coloredcell{+84.54} & \coloredcell{+215.51} & \coloredcell{+253.55} & \coloredcell{+240.55} & \coloredcell{+120.00} & \coloredcell{+157.88} & \coloredcell{+148.49} & 47.79\% & 44.85\% & 45.83\% \\
                & Llama-3.3-70B  & \coloredcell{-29.33} & \coloredcell{-23.20} & \coloredcell{-31.98} & \coloredcell{-21.47} & \coloredcell{-9.52} & \coloredcell{-13.90} & \coloredcell{+25.26} & \coloredcell{+37.46} & \coloredcell{+43.73} & \coloredcell{+0.43} & \coloredcell{+7.67} & \coloredcell{+3.37} & \coloredcell{+5.84} & \coloredcell{+24.93} & \coloredcell{+10.48} & \coloredcell{-53.11} & \coloredcell{-32.67} & \coloredcell{-39.24} & 30.36\% & 33.14\% & 34.55\% \\
                & Qwen-2.5-72B  & \coloredcell{+43.47} & \coloredcell{+39.10} & \coloredcell{+35.98} & \coloredcell{+69.73} & \coloredcell{+57.20} & \coloredcell{+82.92} & \coloredcell{+53.29} & \coloredcell{+52.11} & \coloredcell{+78.27} & \coloredcell{+57.00} & \coloredcell{+62.02} & \coloredcell{+68.42} & \coloredcell{+96.07} & \coloredcell{+109.43} & \coloredcell{+98.97} & \coloredcell{+22.62} & \coloredcell{+49.58} & \coloredcell{+52.81} & 42.87\% & 43.92\% & 48.24\% \\
                & Real  & 5.54 & 5.53 & 5.61 & 1.80 & 1.82 & 1.64 & 0.46 & 0.45 & 0.40 & 1.01 & 1.00 & 0.96 & 890 & 3490 & 1842 & 610 & 2146 & 1157 & 68.54 & 61.49 & 62.81 \\
            \midrule
            \multirow{6}{*}{\shortstack{Weaknesses\\(ICLR 2024)}}
                & GPT-4o  & \coloredcell{-51.57} & \coloredcell{-62.86} & \coloredcell{-67.49} & \coloredcell{-55.73} & \coloredcell{-64.35} & \coloredcell{-68.84} & \coloredcell{-11.74} & \coloredcell{-14.63} & \coloredcell{-14.64} & \coloredcell{-14.59} & \coloredcell{-25.13} & \coloredcell{-26.46} & \coloredcell{-57.26} & \coloredcell{-65.96} & \coloredcell{-69.61} & \coloredcell{-43.46} & \coloredcell{-57.96} & \coloredcell{-64.33} & 92.94\% & 78.14\% & 78.60\% \\
                & Claude-3.5-Sonnet  & \coloredcell{-42.33} & \coloredcell{-61.33} & \coloredcell{-67.39} & \coloredcell{-33.54} & \coloredcell{-54.61} & \coloredcell{-64.97} & \coloredcell{+25.99} & \coloredcell{+13.17} & \coloredcell{+2.53} & \coloredcell{+0.67} & \coloredcell{-19.01} & \coloredcell{-19.08} & \coloredcell{-55.44} & \coloredcell{-72.94} & \coloredcell{-77.67} & \coloredcell{-23.68} & \coloredcell{-42.98} & \coloredcell{-52.03} & 120.31\% & 133.31\% & 143.91\% \\
                & Gemini-2.0-Flash  & \coloredcell{+16.32} & \coloredcell{-14.66} & \coloredcell{-25.02} & \coloredcell{+10.37} & \coloredcell{-19.40} & \coloredcell{-22.21} & \coloredcell{+19.64} & \coloredcell{+2.21} & \coloredcell{+10.81} & \coloredcell{+29.96} & \coloredcell{+9.80} & \coloredcell{+9.27} & \coloredcell{+36.70} & \coloredcell{-2.10} & \coloredcell{-13.74} & \coloredcell{-12.69} & \coloredcell{-34.66} & \coloredcell{-41.86} & 44.87\% & 42.24\% & 45.15\% \\
                & Llama-3.3-70B  & \coloredcell{-66.98} & \coloredcell{-73.49} & \coloredcell{-79.13} & \coloredcell{-79.63} & \coloredcell{-80.48} & \coloredcell{-84.63} & \coloredcell{-38.16} & \coloredcell{-36.14} & \coloredcell{-37.45} & \coloredcell{-40.13} & \coloredcell{-42.89} & \coloredcell{-49.08} & \coloredcell{-55.79} & \coloredcell{-67.56} & \coloredcell{-74.68} & \coloredcell{-82.92} & \coloredcell{-82.86} & \coloredcell{-85.77} & 27.14\% & 33.44\% & 37.65\% \\
                & Qwen-2.5-72B  & \coloredcell{-28.11} & \coloredcell{-55.10} & \coloredcell{-61.72} & \coloredcell{-27.68} & \coloredcell{-55.64} & \coloredcell{-60.28} & \coloredcell{+14.94} & \coloredcell{-4.27} & \coloredcell{+2.36} & \coloredcell{+9.01} & \coloredcell{-14.23} & \coloredcell{-12.82} & \coloredcell{-13.26} & \coloredcell{-49.03} & \coloredcell{-56.18} & \coloredcell{-49.25} & \coloredcell{-64.70} & \coloredcell{-70.01} & 41.10\% & 43.82\% & 45.84\% \\
                & Real  & 7.73 & 10.66 & 12.17 & 2.61 & 3.59 & 3.93 & 0.50 & 0.56 & 0.54 & 1.24 & 1.48 & 1.47 & 1425 & 8842 & 5182 & 1001 & 5595 & 3471 & 70.25\% & 63.28\% & 66.98\% \\
            \midrule
            \multirow{6}{*}{\shortstack{Strengths\\(ICLR 2024)}}
                & GPT-4o  & \coloredcell{+12.14} & \coloredcell{+76.37} & \coloredcell{+142.18} & \coloredcell{+16.93} & \coloredcell{+97.30} & \coloredcell{+193.29} & \coloredcell{+17.28} & \coloredcell{+40.53} & \coloredcell{+71.72} & \coloredcell{+30.51} & \coloredcell{+61.89} & \coloredcell{+112.79} & \coloredcell{+13.13} & \coloredcell{+80.92} & \coloredcell{+162.72} & \coloredcell{+10.06} & \coloredcell{+66.49} & \coloredcell{+103.60} & 46.35\% & 42.39\% & 41.25\% \\
                & Claude-3.5-Sonnet  & \coloredcell{+1.80} & \coloredcell{+50.19} & \coloredcell{+102.36} & \coloredcell{+43.19} & \coloredcell{+103.89} & \coloredcell{+213.77} & \coloredcell{+56.85} & \coloredcell{+70.96} & \coloredcell{+124.83} & \coloredcell{+21.88} & \coloredcell{+56.75} & \coloredcell{+97.80} & \coloredcell{-10.10} & \coloredcell{+37.18} & \coloredcell{+83.42} & \coloredcell{+26.78} & \coloredcell{+78.43} & \coloredcell{+137.95} & 67.19\% & 59.91\% & 69.05\% \\
                & Gemini-2.0-Flash  & \coloredcell{+58.14} & \coloredcell{+138.71} & \coloredcell{+199.86} & \coloredcell{+77.55} & \coloredcell{+156.33} & \coloredcell{+274.25} & \coloredcell{+27.91} & \coloredcell{+34.94} & \coloredcell{+75.50} & \coloredcell{+35.85} & \coloredcell{+71.93} & \coloredcell{+115.56} & \coloredcell{+86.50} & \coloredcell{+180.36} & \coloredcell{+261.52} & \coloredcell{-1.39} & \coloredcell{+47.73} & \coloredcell{+84.02} & 25.19\% & 24.27\% & 27.09\% \\
                & Llama-3.3-70B  & \coloredcell{-43.56} & \coloredcell{-10.29} & \coloredcell{+17.30} & \coloredcell{-60.00} & \coloredcell{-30.37} & \coloredcell{-0.60} & \coloredcell{-34.53} & \coloredcell{-22.69} & \coloredcell{+1.31} & \coloredcell{-24.73} & \coloredcell{-0.63} & \coloredcell{+28.14} & \coloredcell{-23.97} & \coloredcell{+16.62} & \coloredcell{+60.31} & \coloredcell{-84.67} & \coloredcell{-68.73} & \coloredcell{-63.52} & 9.60\% & 12.35\% & 12.11\% \\
                & Qwen-2.5-72B  & \coloredcell{+7.79} & \coloredcell{+46.28} & \coloredcell{+90.10} & \coloredcell{+8.34} & \coloredcell{+52.61} & \coloredcell{+106.35} & \coloredcell{+7.79} & \coloredcell{+27.97} & \coloredcell{+46.91} & \coloredcell{+20.16} & \coloredcell{+51.86} & \coloredcell{+91.62} & \coloredcell{+33.11} & \coloredcell{+78.07} & \coloredcell{+139.05} & \coloredcell{-45.36} & \coloredcell{-22.72} & \coloredcell{-1.86} & 19.56\% & 19.99\% & 21.85\% \\
                & Real  & 6.38 & 4.08 & 3.04 & 2.60 & 1.67 & 1.17 & 0.67 & 0.60 & 0.50 & 1.28 & 1.02 & 0.78 & 1356 & 3784 & 1411 & 646 & 1743 & 751 & 47.64\% & 46.06\% & 53.22\% \\
            \midrule
            \multirow{6}{*}{\shortstack{Summary\\(ICLR 2024)}}
                & GPT-4o  & \coloredcell{+2.59} & \coloredcell{+24.50} & \coloredcell{+38.03} & \coloredcell{+24.60} & \coloredcell{+48.37} & \coloredcell{+73.12} & \coloredcell{+24.29} & \coloredcell{+24.49} & \coloredcell{+34.90} & \coloredcell{+11.76} & \coloredcell{+21.29} & \coloredcell{+26.45} & \coloredcell{+19.83} & \coloredcell{+37.24} & \coloredcell{+59.64} & \coloredcell{-33.30} & \coloredcell{-9.35} & \coloredcell{-9.58} & 26.74\% & 24.87\% & 25.71\% \\
                & Claude-3.5-Sonnet  & \coloredcell{-0.16} & \coloredcell{+17.45} & \coloredcell{+29.74} & \coloredcell{+29.30} & \coloredcell{+51.76} & \coloredcell{+81.05} & \coloredcell{+33.25} & \coloredcell{+35.34} & \coloredcell{+52.81} & \coloredcell{+9.15} & \coloredcell{+18.96} & \coloredcell{+22.65} & \coloredcell{+19.97} & \coloredcell{+31.85} & \coloredcell{+49.87} & \coloredcell{-42.08} & \coloredcell{-20.80} & \coloredcell{-14.61} & 23.19\% & 22.62\% & 25.86\% \\
                & Gemini-2.0-Flash  & \coloredcell{-9.95} & \coloredcell{+7.02} & \coloredcell{+18.77} & \coloredcell{+2.76} & \coloredcell{+23.89} & \coloredcell{+42.35} & \coloredcell{+16.00} & \coloredcell{+19.02} & \coloredcell{+29.82} & \coloredcell{+3.51} & \coloredcell{+12.54} & \coloredcell{+16.12} & \coloredcell{+14.98} & \coloredcell{+25.49} & \coloredcell{+45.89} & \coloredcell{-61.86} & \coloredcell{-42.32} & \coloredcell{-40.98} & 15.94\% & 17.31\% & 18.36\% \\
                & Llama-3.3-70B  & \coloredcell{-27.01} & \coloredcell{-6.11} & \coloredcell{+6.66} & \coloredcell{-21.22} & \coloredcell{+9.45} & \coloredcell{+28.68} & \coloredcell{+8.33} & \coloredcell{+18.68} & \coloredcell{+29.52} & \coloredcell{-4.49} & \coloredcell{+8.18} & \coloredcell{+14.31} & \coloredcell{-1.16} & \coloredcell{+18.96} & \coloredcell{+42.17} & \coloredcell{-80.83} & \coloredcell{-72.67} & \coloredcell{-71.58} & 9.32\% & 8.65\% & 9.07\% \\
                & Qwen-2.5-72B  & \coloredcell{+3.08} & \coloredcell{+16.18} & \coloredcell{+28.64} & \coloredcell{+25.43} & \coloredcell{+46.25} & \coloredcell{+68.39} & \coloredcell{+23.34} & \coloredcell{+30.17} & \coloredcell{+42.28} & \coloredcell{+10.07} & \coloredcell{+19.44} & \coloredcell{+24.95} & \coloredcell{+32.62} & \coloredcell{+39.00} & \coloredcell{+59.61} & \coloredcell{-58.43} & \coloredcell{-44.44} & \coloredcell{-39.61} & 15.06\% & 15.05\% & 17.17\% \\
                & Real  & 9.73 & 8.14 & 7.24 & 4.61 & 3.94 & 3.35 & 0.88 & 0.88 & 0.81 & 1.62 & 1.53 & 1.43 & 2063 & 8007 & 3543 & 991 & 3015 & 1608 & 48.04\% & 37.65\% & 45.39\% \\
            
                \midrule
            \multirow{6}{*}{\shortstack{Questions\\(ICLR 2025)}}
                & GPT-4o  & \coloredcell{-16.66} & \coloredcell{-12.01} & \coloredcell{-12.34} & \coloredcell{+13.05} & \coloredcell{+13.62} & \coloredcell{+16.72} & \coloredcell{+70.82} & \coloredcell{+51.81} & \coloredcell{+60.11} & \coloredcell{+37.18} & \coloredcell{+30.18} & \coloredcell{+33.49} & \coloredcell{+2.64} & \coloredcell{+15.89} & \coloredcell{+24.62} & \coloredcell{-13.29} & \coloredcell{+10.80} & \coloredcell{+7.38} & 56.85\% & 60.66\% & 55.61\% \\
                & Claude-3.5-Sonnet  & \coloredcell{-24.96} & \coloredcell{-16.53} & \coloredcell{-12.13} & \coloredcell{-25.25} & \coloredcell{-21.99} & \coloredcell{-16.79} & \coloredcell{+27.55} & \coloredcell{+9.81} & \coloredcell{+13.58} & \coloredcell{+36.71} & \coloredcell{+32.31} & \coloredcell{+40.75} & \coloredcell{-7.24} & \coloredcell{+8.56} & \coloredcell{+24.25} & \coloredcell{-21.90} & \coloredcell{+4.12} & \coloredcell{+9.21} & 56.66\% & 60.85\% & 56.73\% \\
                & Gemini-2.0-Flash  & \coloredcell{+110.55} & \coloredcell{+111.94} & \coloredcell{+116.65} & \coloredcell{+133.90} & \coloredcell{+111.21} & \coloredcell{+133.38} & \coloredcell{+52.15} & \coloredcell{+27.98} & \coloredcell{+38.03} & \coloredcell{+78.26} & \coloredcell{+64.94} & \coloredcell{+68.21} & \coloredcell{+188.50} & \coloredcell{+215.13} & \coloredcell{+240.97} & \coloredcell{+77.59} & \coloredcell{+109.03} & \coloredcell{+114.50} & 41.42\% & 42.08\% & 40.60\% \\
                & Llama-3.3-70B  & \coloredcell{-39.53} & \coloredcell{-38.69} & \coloredcell{-35.50} & \coloredcell{-20.17} & \coloredcell{-30.42} & \coloredcell{-25.56} & \coloredcell{+53.38} & \coloredcell{+20.64} & \coloredcell{+23.65} & \coloredcell{-1.48} & \coloredcell{-9.06} & \coloredcell{-3.06} & \coloredcell{-12.01} & \coloredcell{-1.09} & \coloredcell{+11.18} & \coloredcell{-59.37} & \coloredcell{-53.50} & \coloredcell{-50.92} & 31.07\% & 29.83\% & 28.49\% \\
                & Qwen-2.5-72B  & \coloredcell{+21.80} & \coloredcell{+28.04} & \coloredcell{+33.28} & \coloredcell{+57.97} & \coloredcell{+45.19} & \coloredcell{+73.90} & \coloredcell{+79.06} & \coloredcell{+43.88} & \coloredcell{+66.08} & \coloredcell{+62.02} & \coloredcell{+51.76} & \coloredcell{+56.50} & \coloredcell{+68.91} & \coloredcell{+89.14} & \coloredcell{+109.49} & \coloredcell{-3.04} & \coloredcell{+23.41} & \coloredcell{+34.28} & 38.63\% & 41.40\% & 41.37\% \\
                & Real  & 6.13 & 5.92 & 5.79 & 1.84 & 1.86 & 1.75 & 0.41 & 0.46 & 0.43 & 1.00 & 1.08 & 1.05 & 1174 & 3562 & 2961 & 790.00 & 2260 & 1911 & 67.29\% & 63.45\% & 64.54\% \\
            \midrule
            \multirow{6}{*}{\shortstack{Weakness\\(ICLR 2025)}}
                & GPT-4o  & \coloredcell{-59.42} & \coloredcell{-65.46} & \coloredcell{-71.45} & \coloredcell{-64.34} & \coloredcell{-68.81} & \coloredcell{-73.91} & \coloredcell{-13.85} & \coloredcell{-22.77} & \coloredcell{-19.22} & \coloredcell{-16.36} & \coloredcell{-28.62} & \coloredcell{-30.65} & \coloredcell{-63.72} & \coloredcell{-68.16} & \coloredcell{-72.41} & \coloredcell{-53.08} & \coloredcell{-61.40} & \coloredcell{-70.05} & 87.90\% & 80.59\% & 73.69\% \\
                & Claude-3.5-Sonnet  & \coloredcell{-52.91} & \coloredcell{-63.14} & \coloredcell{-70.40} & \coloredcell{-38.86} & \coloredcell{-59.91} & \coloredcell{-69.04} & \coloredcell{+37.50} & \coloredcell{+2.60} & \coloredcell{+0.58} & \coloredcell{-5.54} & \coloredcell{-21.89} & \coloredcell{-24.12} & \coloredcell{-59.68} & \coloredcell{-71.55} & \coloredcell{-78.66} & \coloredcell{-42.95} & \coloredcell{-50.49} & \coloredcell{-58.22} & 96.16\% & 115.68\% & 132.91\% \\
                & Gemini-2.0-Flash  & \coloredcell{-1.60} & \coloredcell{-20.50} & \coloredcell{-33.96} & \coloredcell{-4.58} & \coloredcell{-26.58} & \coloredcell{-37.00} & \coloredcell{+20.41} & \coloredcell{-4.34} & \coloredcell{-1.29} & \coloredcell{+26.26} & \coloredcell{+6.06} & \coloredcell{-0.39} & \coloredcell{+19.31} & \coloredcell{-5.36} & \coloredcell{-21.17} & \coloredcell{-32.38} & \coloredcell{-43.28} & \coloredcell{-52.79} & 38.52\% & 39.83\% & 40.65\% \\
                & Llama-3.3-70B  & \coloredcell{-73.08} & \coloredcell{-78.11} & \coloredcell{-81.84} & \coloredcell{-77.54} & \coloredcell{-83.87} & \coloredcell{-86.86} & \coloredcell{-24.35} & \coloredcell{-39.61} & \coloredcell{-41.80} & \coloredcell{-43.24} & \coloredcell{-50.43} & \coloredcell{-53.68} & \coloredcell{-63.02} & \coloredcell{-72.10} & \coloredcell{-75.96} & \coloredcell{-87.89} & \coloredcell{-87.14} & \coloredcell{-90.50} & 22.27\% & 30.63\% & 26.83\% \\
                & Qwen-2.5-72B  & \coloredcell{-47.45} & \coloredcell{-57.88} & \coloredcell{-66.15} & \coloredcell{-47.48} & \coloredcell{-60.46} & \coloredcell{-67.10} & \coloredcell{+11.94} & \coloredcell{-9.77} & \coloredcell{-7.78} & \coloredcell{+1.46} & \coloredcell{-14.09} & \coloredcell{-21.87} & \coloredcell{-33.43} & \coloredcell{-50.48} & \coloredcell{-58.79} & \coloredcell{-68.06} & \coloredcell{-69.01} & \coloredcell{-76.99} & 32.61\% & 41.60\% & 37.90\% \\
                & Real  & 9.12 & 11.20 & 13.68 & 2.96 & 3.85 & 4.57 & 0.47 & 0.59 & 0.57 & 1.28 & 1.52 & 1.60 & 2004 & 8496 & 9271 & 1362 & 5647 & 6293 & 67.96\% & 66.47\% & 67.88\% \\
            \midrule
            \multirow{6}{*}{\shortstack{Strengths\\(ICLR 2025)}}
    
                & GPT-4o  & \coloredcell{+15.74} & \coloredcell{+63.44} & \coloredcell{+109.96} & \coloredcell{+25.47} & \coloredcell{+71.53} & \coloredcell{+146.12} & \coloredcell{+22.84} & \coloredcell{+34.61} & \coloredcell{+56.85} & \coloredcell{+31.75} & \coloredcell{+55.74} & \coloredcell{+84.68} & \coloredcell{+20.80} & \coloredcell{+70.33} & \coloredcell{+126.53} & \coloredcell{+5.84} & \coloredcell{+48.87} & \coloredcell{+76.36} & 44.76\% & 41.34\% & 38.40\% \\
                & Claude-3.5-Sonnet  & \coloredcell{+1.88} & \coloredcell{+37.65} & \coloredcell{+71.33} & \coloredcell{+39.95} & \coloredcell{+77.55} & \coloredcell{+156.93} & \coloredcell{+57.76} & \coloredcell{+66.17} & \coloredcell{+100.96} & \coloredcell{+26.20} & \coloredcell{+42.89} & \coloredcell{+69.83} & \coloredcell{-2.64} & \coloredcell{+26.93} & \coloredcell{+56.13} & \coloredcell{+10.74} & \coloredcell{+60.31} & \coloredcell{+102.15} & 58.11\% & 59.73\% & 63.85\% \\
                & Gemini-2.0-Flash  & \coloredcell{+64.30} & \coloredcell{+115.58} & \coloredcell{+167.86} & \coloredcell{+77.80} & \coloredcell{+126.18} & \coloredcell{+214.25} & \coloredcell{+25.51} & \coloredcell{+34.78} & \coloredcell{+58.71} & \coloredcell{+44.74} & \coloredcell{+63.55} & \coloredcell{+92.16} & \coloredcell{+100.61} & \coloredcell{+157.62} & \coloredcell{+225.74} & \coloredcell{-6.76} & \coloredcell{+26.70} & \coloredcell{+50.50} & 23.74\% & 23.26\% & 22.79\% \\
                & Llama-3.3-70B  & \coloredcell{-39.78} & \coloredcell{-20.48} & \coloredcell{+2.41} & \coloredcell{-48.71} & \coloredcell{-37.28} & \coloredcell{-10.75} & \coloredcell{-18.73} & \coloredcell{-18.32} & \coloredcell{-6.64} & \coloredcell{-17.12} & \coloredcell{-5.44} & \coloredcell{+11.47} & \coloredcell{-17.48} & \coloredcell{+6.09} & \coloredcell{+39.44} & \coloredcell{-83.42} & \coloredcell{-76.67} & \coloredcell{-72.68} & 10.26\% & 10.40\% & 9.66\% \\
                & Qwen-2.5-72B  & \coloredcell{+0.36} & \coloredcell{+33.39} & \coloredcell{+70.60} & \coloredcell{+1.17} & \coloredcell{+30.73} & \coloredcell{+83.40} & \coloredcell{+17.35} & \coloredcell{+20.23} & \coloredcell{+38.42} & \coloredcell{+20.90} & \coloredcell{+40.95} & \coloredcell{+68.62} & \coloredcell{+22.90} & \coloredcell{+65.33} & \coloredcell{+117.11} & \coloredcell{-54.77} & \coloredcell{-34.16} & \coloredcell{-23.71} & 18.80\% & 18.83\% & 17.33\% \\
                & Real  & 6.04 & 4.46 & 3.47 & 2.32 & 1.93 & 1.38 & 0.61 & 0.62 & 0.54 & 1.21 & 1.07 & 0.90 & 1476 & 3825 & 2642 & 754 & 1809 & 1303 & 51.08\% & 47.29\% & 49.32\% \\
            \midrule
            \multirow{6}{*}{\shortstack{Summary\\(ICLR 2025)}}
    
                & GPT-4o  & \coloredcell{-2.44} & \coloredcell{+21.40} & \coloredcell{+36.00} & \coloredcell{+6.65} & \coloredcell{+39.86} & \coloredcell{+64.30} & \coloredcell{+13.31} & \coloredcell{+20.55} & \coloredcell{+28.33} & \coloredcell{+10.43} & \coloredcell{+15.97} & \coloredcell{+23.65} & \coloredcell{+13.12} & \coloredcell{+35.89} & \coloredcell{+55.97} & \coloredcell{-38.05} & \coloredcell{-15.37} & \coloredcell{-12.96} & 23.94\% & 24.55\% & 22.75\% \\
                & Claude-3.5-Sonnet  & \coloredcell{-2.53} & \coloredcell{+13.25} & \coloredcell{+23.68} & \coloredcell{+15.49} & \coloredcell{+35.35} & \coloredcell{+58.01} & \coloredcell{+24.81} & \coloredcell{+26.43} & \coloredcell{+37.54} & \coloredcell{+9.54} & \coloredcell{+13.53} & \coloredcell{+21.22} & \coloredcell{+18.39} & \coloredcell{+29.41} & \coloredcell{+43.74} & \coloredcell{-50.36} & \coloredcell{-27.73} & \coloredcell{-25.51} & 18.33\% & 22.02\% & 21.13\% \\
                & Gemini-2.0-Flash  & \coloredcell{-9.24} & \coloredcell{+5.71} & \coloredcell{+14.10} & \coloredcell{-3.71} & \coloredcell{+21.45} & \coloredcell{+28.21} & \coloredcell{+9.96} & \coloredcell{+19.48} & \coloredcell{+19.37} & \coloredcell{+4.75} & \coloredcell{+6.61} & \coloredcell{+14.77} & \coloredcell{+14.56} & \coloredcell{+27.24} & \coloredcell{+40.41} & \coloredcell{-63.69} & \coloredcell{-48.90} & \coloredcell{-50.42} & 13.86\% & 15.83\% & 14.40\% \\
                & Llama-3.3-70B  & \coloredcell{-28.14} & \coloredcell{-12.73} & \coloredcell{+0.53} & \coloredcell{-22.91} & \coloredcell{-2.29} & \coloredcell{+19.42} & \coloredcell{+10.20} & \coloredcell{+15.45} & \coloredcell{+25.76} & \coloredcell{+0.08} & \coloredcell{+1.93} & \coloredcell{+11.36} & \coloredcell{-2.95} & \coloredcell{+13.50} & \coloredcell{+31.13} & \coloredcell{-85.77} & \coloredcell{-79.26} & \coloredcell{-74.53} & 6.41\% & 7.20\% & 7.92\% \\
                & Qwen-2.5-72B  & \coloredcell{-6.86} & \coloredcell{+14.97} & \coloredcell{+27.99} & \coloredcell{+8.30} & \coloredcell{+39.17} & \coloredcell{+63.96} & \coloredcell{+20.83} & \coloredcell{+26.35} & \coloredcell{+36.29} & \coloredcell{+6.91} & \coloredcell{+12.76} & \coloredcell{+22.17} & \coloredcell{+19.51} & \coloredcell{+41.15} & \coloredcell{+59.36} & \coloredcell{-67.15} & \coloredcell{-51.44} & \coloredcell{-48.93} & 12.02\% & 13.56\% & 13.07\% \\
                & Real  & 9.76 & 8.29 & 7.51 & 4.60 & 4.14 & 3.71 & 0.86 & 0.91 & 0.87 & 1.57 & 1.57 & 1.48 & 2507 & 7511 & 6075 & 1096 & 2961 & 2477 & 43.72\% & 39.42\% & 40.77\% \\
            \midrule
            \multirow{6}{*}{\shortstack{Questions\\(NeurIPS 2023)}}
                & GPT-4o  & \coloredcell{+5.20} & \coloredcell{+5.65} & \coloredcell{+2.36} & \coloredcell{+45.18} & \coloredcell{+44.93} & \coloredcell{+24.87} & \coloredcell{+77.27} & \coloredcell{+70.76} & \coloredcell{+71.51} & \coloredcell{+57.51} & \coloredcell{+46.49} & \coloredcell{+44.72} & \coloredcell{+13.79} & \coloredcell{+26.51} & \coloredcell{+27.64} & \coloredcell{+26.46} & \coloredcell{+37.30} & \coloredcell{+17.52} & 69.78\% & 68.53\% & 78.35\% \\
                & Claude-3.5-Sonnet  & \coloredcell{-0.31} & \coloredcell{-9.00} & \coloredcell{-6.48} & \coloredcell{+13.95} & \coloredcell{-13.59} & \coloredcell{-19.17} & \coloredcell{+40.82} & \coloredcell{+15.46} & \coloredcell{+25.71} & \coloredcell{+57.49} & \coloredcell{+31.17} & \coloredcell{+44.54} & \coloredcell{+9.14} & \coloredcell{+21.96} & \coloredcell{+21.09} & \coloredcell{+18.78} & \coloredcell{+2.16} & \coloredcell{+2.14} & 68.34\% & 52.89\% & 71.77\% \\
                & Gemini-2.0-Flash  & \coloredcell{+126.94} & \coloredcell{+119.25} & \coloredcell{+101.77} & \coloredcell{+146.51} & \coloredcell{+133.18} & \coloredcell{+99.48} & \coloredcell{+47.95} & \coloredcell{+44.55} & \coloredcell{+43.26} & \coloredcell{+88.43} & \coloredcell{+65.93} & \coloredcell{+59.14} & \coloredcell{+211.63} & \coloredcell{+222.07} & \coloredcell{+221.45} & \coloredcell{+77.51} & \coloredcell{+104.14} & \coloredcell{+49.57} & 35.77\% & 40.02\% & 39.59\% \\
                & Llama-3.3-70B  & \coloredcell{-32.55} & \coloredcell{-32.15} & \coloredcell{-33.20} & \coloredcell{-37.87} & \coloredcell{-31.34} & \coloredcell{-29.53} & \coloredcell{+1.21} & \coloredcell{+13.78} & \coloredcell{+25.71} & \coloredcell{+1.92} & \coloredcell{+0.37} & \coloredcell{-3.32} & \coloredcell{-6.48} & \coloredcell{+6.37} & \coloredcell{+14.18} & \coloredcell{-53.17} & \coloredcell{-49.91} & \coloredcell{-53.85} & 31.44\% & 29.73\% & 34.39\% \\
                & Qwen-2.5-72B  & \coloredcell{+41.43} & \coloredcell{+40.10} & \coloredcell{+33.40} & \coloredcell{+73.42} & \coloredcell{+69.59} & \coloredcell{+29.02} & \coloredcell{+61.67} & \coloredcell{+61.22} & \coloredcell{+39.19} & \coloredcell{+75.51} & \coloredcell{+56.32} & \coloredcell{+51.25} & \coloredcell{+89.70} & \coloredcell{+100.68} & \coloredcell{+105.45} & \coloredcell{+17.46} & \coloredcell{+29.73} & \coloredcell{-0.43} & 38.88\% & 40.82\% & 41.24\% \\
                & Real  & 5.44 & 5.78 & 6.13 & 1.67 & 1.75 & 2.33 & 0.42 & 0.41 & 0.48 & 0.92 & 1.05 & 1.13 & 602 & 879 & 275 & 378 & 555 & 234 & 62.79\% & 63.14\% & 85.09\% \\
            \midrule
            \multirow{6}{*}{\shortstack{Weaknesses\\(NeurIPS 2023)}}
                & GPT-4o  & \coloredcell{-17.34} & \coloredcell{-21.61} & \coloredcell{-42.17} & \coloredcell{+5.45} & \coloredcell{-14.59} & \coloredcell{-45.28} & \coloredcell{+78.35} & \coloredcell{+25.08} & \coloredcell{+5.26} & \coloredcell{+32.04} & \coloredcell{+20.90} & \coloredcell{+14.17} & \coloredcell{-29.26} & \coloredcell{-30.09} & \coloredcell{-46.10} & \coloredcell{-0.17} & \coloredcell{-9.34} & \coloredcell{-36.38} & 98.05\% & 89.62\% & 80.16\% \\
                & Claude-3.5-Sonnet  & \coloredcell{-36.85} & \coloredcell{-48.28} & \coloredcell{-62.22} & \coloredcell{-18.18} & \coloredcell{-46.19} & \coloredcell{-61.74} & \coloredcell{+62.29} & \coloredcell{+9.54} & \coloredcell{+3.71} & \coloredcell{+8.45} & \coloredcell{-2.67} & \coloredcell{-1.51} & \coloredcell{-44.35} & \coloredcell{-56.27} & \coloredcell{-73.99} & \coloredcell{-26.04} & \coloredcell{-36.71} & \coloredcell{-44.89} & 92.34\% & 100.00\% & 143.89\% \\
                & Gemini-2.0-Flash  & \coloredcell{+37.87} & \coloredcell{+19.44} & \coloredcell{-5.51} & \coloredcell{+72.21} & \coloredcell{+12.04} & \coloredcell{-6.30} & \coloredcell{+89.56} & \coloredcell{+12.76} & \coloredcell{+18.56} & \coloredcell{+55.35} & \coloredcell{+34.72} & \coloredcell{+28.54} & \coloredcell{+66.24} & \coloredcell{+44.17} & \coloredcell{+10.40} & \coloredcell{-2.99} & \coloredcell{-16.34} & \coloredcell{-28.94} & 40.54\% & 40.10\% & 43.72\% \\
                & Llama-3.3-70B  & \coloredcell{-67.91} & \coloredcell{-67.68} & \coloredcell{-76.76} & \coloredcell{-77.14} & \coloredcell{-78.12} & \coloredcell{-86.20} & \coloredcell{-28.48} & \coloredcell{-34.17} & \coloredcell{-42.44} & \coloredcell{-44.29} & \coloredcell{-36.17} & \coloredcell{-42.24} & \coloredcell{-55.41} & \coloredcell{-55.89} & \coloredcell{-70.09} & \coloredcell{-85.90} & \coloredcell{-84.74} & \coloredcell{-86.60} & 21.96\% & 23.91\% & 30.43\% \\
                & Qwen-2.5-72B  & \coloredcell{+3.87} & \coloredcell{-5.76} & \coloredcell{-24.78} & \coloredcell{+35.58} & \coloredcell{-7.96} & \coloredcell{-23.00} & \coloredcell{+89.59} & \coloredcell{+17.24} & \coloredcell{+18.10} & \coloredcell{+48.64} & \coloredcell{+31.09} & \coloredcell{+23.58} & \coloredcell{+31.45} & \coloredcell{+18.05} & \coloredcell{-6.36} & \coloredcell{-35.82} & \coloredcell{-40.22} & \coloredcell{-51.91} & 33.92\% & 35.00\% & 34.88\% \\
                & Real  & 7.14 & 8.71 & 11.62 & 1.87 & 2.89 & 4.13 & 0.32 & 0.50 & 0.55 & 1.08 & 1.26 & 1.37 & 868 & 1612 & 692 & 603 & 1114 & 470 & 69.47\% & 69.11\% & 67.92\% \\
            \midrule
            \multirow{6}{*}{\shortstack{Strengths\\(NeurIPS 2023)}}
                & GPT-4o  & \coloredcell{+69.84} & \coloredcell{+111.97} & \coloredcell{+164.66} & \coloredcell{+80.45} & \coloredcell{+128.59} & \coloredcell{+190.41} & \coloredcell{+30.78} & \coloredcell{+32.62} & \coloredcell{+42.53} & \coloredcell{+47.67} & \coloredcell{+74.10} & \coloredcell{+77.50} & \coloredcell{+60.99} & \coloredcell{+109.84} & \coloredcell{+150.19} & \coloredcell{+89.41} & \coloredcell{+116.29} & \coloredcell{+195.12} & 53.27\% & 50.81\% & 56.02\% \\
                & Claude-3.5-Sonnet  & \coloredcell{+14.30} & \coloredcell{+23.60} & \coloredcell{+40.58} & \coloredcell{+47.52} & \coloredcell{+50.82} & \coloredcell{+69.18} & \coloredcell{+47.28} & \coloredcell{+41.14} & \coloredcell{+41.73} & \coloredcell{+23.20} & \coloredcell{+41.93} & \coloredcell{+45.91} & \coloredcell{+5.06} & \coloredcell{+15.26} & \coloredcell{+30.12} & \coloredcell{+34.71} & \coloredcell{+40.53} & \coloredcell{+62.60} & 58.05\% & 60.10\% & 59.35\% \\
                & Gemini-2.0-Flash  & \coloredcell{+111.82} & \coloredcell{+152.25} & \coloredcell{+207.33} & \coloredcell{+155.69} & \coloredcell{+176.63} & \coloredcell{+263.70} & \coloredcell{+54.85} & \coloredcell{+36.07} & \coloredcell{+55.89} & \coloredcell{+51.12} & \coloredcell{+75.08} & \coloredcell{+88.01} & \coloredcell{+153.66} & \coloredcell{+203.71} & \coloredcell{+256.76} & \coloredcell{+19.41} & \coloredcell{+47.86} & \coloredcell{+103.25} & 21.31\% & 24.00\% & 27.06\% \\
                & Llama-3.3-70B  & \coloredcell{-33.36} & \coloredcell{-15.40} & \coloredcell{+2.62} & \coloredcell{-48.27} & \coloredcell{-32.68} & \coloredcell{-9.59} & \coloredcell{-15.69} & \coloredcell{-17.38} & \coloredcell{-4.45} & \coloredcell{-14.61} & \coloredcell{+7.27} & \coloredcell{+8.74} & \coloredcell{-13.85} & \coloredcell{+13.15} & \coloredcell{+37.84} & \coloredcell{-76.47} & \coloredcell{-73.32} & \coloredcell{-71.54} & 12.36\% & 11.62\% & 9.80\% \\
                & Qwen-2.5-72B  & \coloredcell{+60.31} & \coloredcell{+87.29} & \coloredcell{+132.46} & \coloredcell{+62.13} & \coloredcell{+87.75} & \coloredcell{+143.15} & \coloredcell{+22.93} & \coloredcell{+21.66} & \coloredcell{+35.15} & \coloredcell{+39.58} & \coloredcell{+65.29} & \coloredcell{+72.54} & \coloredcell{+95.21} & \coloredcell{+137.35} & \coloredcell{+187.26} & \coloredcell{-16.76} & \coloredcell{-14.26} & \coloredcell{+17.07} & 19.30\% & 17.81\% & 19.35\% \\
                & Real  & 5.30 & 4.75 & 3.82 & 1.96 & 1.96 & 1.46 & 0.55 & 0.62 & 0.55 & 1.17 & 1.04 & 1.01 & 751 & 996 & 259 & 340 & 491 & 123 & 45.27\% & 49.30\% & 47.49\% \\
            \midrule
            \multirow{6}{*}{\shortstack{Summary\\(NeurIPS 2023)}}
                & GPT-4o  & \coloredcell{-9.38} & \coloredcell{+9.15} & \coloredcell{+31.26} & \coloredcell{+6.83} & \coloredcell{+38.75} & \coloredcell{+46.46} & \coloredcell{+16.61} & \coloredcell{+32.50} & \coloredcell{+21.93} & \coloredcell{+6.42} & \coloredcell{+18.02} & \coloredcell{+21.64} & \coloredcell{+2.69} & \coloredcell{+24.82} & \coloredcell{+53.12} & \coloredcell{-39.10} & \coloredcell{-24.83} & \coloredcell{-16.53} & 24.09\% & 27.76\% & 24.94\% \\
                & Claude-3.5-Sonnet  & \coloredcell{-8.40} & \coloredcell{+5.91} & \coloredcell{+24.51} & \coloredcell{+13.78} & \coloredcell{+46.39} & \coloredcell{+44.70} & \coloredcell{+24.28} & \coloredcell{+47.27} & \coloredcell{+29.99} & \coloredcell{+1.57} & \coloredcell{+17.70} & \coloredcell{+21.16} & \coloredcell{+7.87} & \coloredcell{+25.23} & \coloredcell{+48.77} & \coloredcell{-48.47} & \coloredcell{-36.00} & \coloredcell{-28.51} & 19.40\% & 23.56\% & 21.98\% \\
                & Gemini-2.0-Flash  & \coloredcell{-23.47} & \coloredcell{-5.38} & \coloredcell{+11.28} & \coloredcell{-14.12} & \coloredcell{+21.36} & \coloredcell{+14.90} & \coloredcell{+13.77} & \coloredcell{+31.84} & \coloredcell{+12.15} & \coloredcell{-0.87} & \coloredcell{+5.16} & \coloredcell{+11.66} & \coloredcell{-5.31} & \coloredcell{+16.34} & \coloredcell{+41.02} & \coloredcell{-68.17} & \coloredcell{-52.48} & \coloredcell{-53.72} & 13.65\% & 18.83\% & 15.01\% \\
                & Llama-3.3-70B  & \coloredcell{-32.29} & \coloredcell{-21.05} & \coloredcell{-5.97} & \coloredcell{-20.84} & \coloredcell{-9.10} & \coloredcell{-5.56} & \coloredcell{+16.12} & \coloredcell{+17.18} & \coloredcell{+9.44} & \coloredcell{-6.36} & \coloredcell{+2.03} & \coloredcell{+4.88} & \coloredcell{-14.30} & \coloredcell{+5.31} & \coloredcell{+25.90} & \coloredcell{-76.58} & \coloredcell{-78.22} & \coloredcell{-75.62} & 11.10\% & 9.54\% & 8.86\% \\
                & Qwen-2.5-72B  & \coloredcell{-14.51} & \coloredcell{+0.50} & \coloredcell{+14.14} & \coloredcell{+1.59} & \coloredcell{+37.12} & \coloredcell{+34.34} & \coloredcell{+20.89} & \coloredcell{+42.54} & \coloredcell{+30.13} & \coloredcell{+7.24} & \coloredcell{+12.46} & \coloredcell{+16.93} & \coloredcell{+5.64} & \coloredcell{+24.40} & \coloredcell{+39.32} & \coloredcell{-64.14} & \coloredcell{-51.35} & \coloredcell{-40.91} & 13.79\% & 18.03\% & 19.40\% \\
                & Real  & 10.40 & 8.97 & 7.71 & 4.26 & 3.93 & 3.96 & 0.77 & 0.79 & 0.89 & 1.61 & 1.54 & 1.57 & 1524 & 1922 & 529 & 619 & 886 & 242 & 40.62\% & 46.10\% & 45.75\% \\
                
            \midrule
            \multirow{6}{*}{\shortstack{Questions\\(NeurIPS 2024)}}
                & GPT-4o  & \coloredcell{-7.74} & \coloredcell{-1.94} & \coloredcell{-14.40} & \coloredcell{+5.81} & \coloredcell{+17.58} & \coloredcell{+5.33} & \coloredcell{+37.58} & \coloredcell{+51.55} & \coloredcell{+41.74} & \coloredcell{+37.10} & \coloredcell{+32.05} & \coloredcell{+41.48} & \coloredcell{+10.01} & \coloredcell{+17.84} & \coloredcell{+7.48} & \coloredcell{+8.65} & \coloredcell{+26.45} & \coloredcell{+11.83} & 63.72\% & 63.79\% & 60.29\% \\
                & Claude-3.5-Sonnet  & \coloredcell{-13.30} & \coloredcell{-3.55} & \coloredcell{-16.77} & \coloredcell{-14.90} & \coloredcell{+4.97} & \coloredcell{-10.67} & \coloredcell{+20.38} & \coloredcell{+42.65} & \coloredcell{+32.76} & \coloredcell{+38.74} & \coloredcell{+37.19} & \coloredcell{+41.61} & \coloredcell{+4.01} & \coloredcell{+19.61} & \coloredcell{+8.10} & \coloredcell{+5.54} & \coloredcell{+14.12} & \coloredcell{+5.38} & 65.47\% & 56.72\% & 56.48\% \\
                & Gemini-2.0-Flash  & \coloredcell{+105.39} & \coloredcell{+109.42} & \coloredcell{+84.81} & \coloredcell{+104.55} & \coloredcell{+114.74} & \coloredcell{+130.00} & \coloredcell{+30.35} & \coloredcell{+38.19} & \coloredcell{+70.00} & \coloredcell{+70.02} & \coloredcell{+63.35} & \coloredcell{+62.99} & \coloredcell{+184.84} & \coloredcell{+198.59} & \coloredcell{+163.55} & \coloredcell{+85.37} & \coloredcell{+102.38} & \coloredcell{+95.16} & 41.99\% & 40.30\% & 42.91\% \\
                & Llama-3.3-70B  & \coloredcell{-43.04} & \coloredcell{-35.03} & \coloredcell{-33.93} & \coloredcell{-41.67} & \coloredcell{-22.56} & \coloredcell{-14.00} & \coloredcell{+14.06} & \coloredcell{+28.39} & \coloredcell{+56.57} & \coloredcell{-9.56} & \coloredcell{-12.18} & \coloredcell{+7.55} & \coloredcell{-11.16} & \coloredcell{-0.80} & \coloredcell{-3.74} & \coloredcell{-62.31} & \coloredcell{-52.75} & \coloredcell{-45.70} & 27.38\% & 28.32\% & 32.69\% \\
                & Qwen-2.5-72B  & \coloredcell{+27.83} & \coloredcell{+38.03} & \coloredcell{+33.73} & \coloredcell{+40.40} & \coloredcell{+52.58} & \coloredcell{+74.67} & \coloredcell{+42.78} & \coloredcell{+50.33} & \coloredcell{+70.31} & \coloredcell{+57.26} & \coloredcell{+47.60} & \coloredcell{+65.07} & \coloredcell{+78.40} & \coloredcell{+93.20} & \coloredcell{+89.10} & \coloredcell{+15.96} & \coloredcell{+31.80} & \coloredcell{+36.56} & 41.94\% & 40.56\% & 41.85\% \\
                & Real  & 5.96 & 5.62 & 6.42 & 2.05 & 1.75 & 1.90 & 0.48 & 0.42 & 0.42 & 1.00 & 1.07 & 1.06 & 699 & 1132 & 321 & 451 & 673 & 186 & 64.52\% & 59.45\% & 57.94\% \\
            \midrule
            \multirow{6}{*}{\shortstack{Weaknesses\\(NeurIPS 2024)}}
                & GPT-4o  & \coloredcell{-1.82} & \coloredcell{-26.48} & \coloredcell{-45.64} & \coloredcell{+9.69} & \coloredcell{-22.46} & \coloredcell{-42.23} & \coloredcell{+27.46} & \coloredcell{+10.49} & \coloredcell{+4.63} & \coloredcell{+29.36} & \coloredcell{+8.44} & \coloredcell{-1.50} & \coloredcell{-9.30} & \coloredcell{-36.06} & \coloredcell{-47.32} & \coloredcell{+8.43} & \coloredcell{-10.49} & \coloredcell{-43.26} & 87.26\% & 83.82\% & 75.87\% \\
                & Claude-3.5-Sonnet  & \coloredcell{-31.01} & \coloredcell{-49.94} & \coloredcell{-64.24} & \coloredcell{-18.89} & \coloredcell{-45.08} & \coloredcell{-55.43} & \coloredcell{+23.59} & \coloredcell{+5.53} & \coloredcell{+14.85} & \coloredcell{+8.94} & \coloredcell{-11.39} & \coloredcell{-25.86} & \coloredcell{-38.69} & \coloredcell{-60.33} & \coloredcell{-77.79} & \coloredcell{-20.48} & \coloredcell{-32.60} & \coloredcell{-45.00} & 94.67\% & 101.71\% & 174.48\% \\
                & Gemini-2.0-Flash  & \coloredcell{+71.17} & \coloredcell{+12.04} & \coloredcell{-8.27} & \coloredcell{+94.67} & \coloredcell{+8.07} & \coloredcell{-4.40} & \coloredcell{+46.57} & \coloredcell{+12.28} & \coloredcell{+0.27} & \coloredcell{+70.55} & \coloredcell{+26.61} & \coloredcell{+15.45} & \coloredcell{+109.30} & \coloredcell{+26.30} & \coloredcell{+6.74} & \coloredcell{+18.93} & \coloredcell{-11.77} & \coloredcell{-29.57} & 41.48\% & 41.82\% & 46.48\% \\
                & Llama-3.3-70B  & \coloredcell{-61.44} & \coloredcell{-72.19} & \coloredcell{-76.46} & \coloredcell{-70.94} & \coloredcell{-83.10} & \coloredcell{-78.89} & \coloredcell{-29.65} & \coloredcell{-47.77} & \coloredcell{-36.56} & \coloredcell{-32.96} & \coloredcell{-48.75} & \coloredcell{-47.92} & \coloredcell{-46.98} & \coloredcell{-64.80} & \coloredcell{-69.83} & \coloredcell{-81.24} & \coloredcell{-84.53} & \coloredcell{-85.87} & 25.83\% & 26.32\% & 32.99\% \\
                & Qwen-2.5-72B  & \coloredcell{+37.69} & \coloredcell{-8.59} & \coloredcell{-23.27} & \coloredcell{+64.65} & \coloredcell{-4.71} & \coloredcell{-20.82} & \coloredcell{+51.04} & \coloredcell{+17.44} & \coloredcell{+5.66} & \coloredcell{+56.67} & \coloredcell{+20.77} & \coloredcell{+12.13} & \coloredcell{+69.47} & \coloredcell{+3.84} & \coloredcell{-7.20} & \coloredcell{-5.85} & \coloredcell{-29.36} & \coloredcell{-46.09} & 40.55\% & 40.73\% & 40.92\% \\
                & Real  & 5.96 & 8.96 & 11.02 & 1.79 & 3.01 & 3.38 & 0.43 & 0.54 & 0.56 & 1.02 & 1.37 & 1.50 & 796 & 2213 & 653 & 581 & 1325 & 460 & 72.99\% & 59.87\% & 70.44\% \\
            \midrule
            \multirow{6}{*}{\shortstack{Strengths\\(NeurIPS 2024)}}
                & GPT-4o  & \coloredcell{+79.06} & \coloredcell{+108.97} & \coloredcell{+133.88} & \coloredcell{+88.53} & \coloredcell{+139.51} & \coloredcell{+134.03} & \coloredcell{+49.78} & \coloredcell{+37.99} & \coloredcell{+29.55} & \coloredcell{+62.08} & \coloredcell{+56.48} & \coloredcell{+61.61} & \coloredcell{+77.16} & \coloredcell{+112.29} & \coloredcell{+148.56} & \coloredcell{+83.42} & \coloredcell{+101.83} & \coloredcell{+106.12} & 45.09\% & 44.16\% & 43.85\% \\
                & Claude-3.5-Sonnet  & \coloredcell{-5.69} & \coloredcell{+6.91} & \coloredcell{+11.06} & \coloredcell{+8.65} & \coloredcell{+41.02} & \coloredcell{+19.90} & \coloredcell{+47.68} & \coloredcell{+43.08} & \coloredcell{+23.98} & \coloredcell{+16.80} & \coloredcell{+15.71} & \coloredcell{+18.02} & \coloredcell{-9.23} & \coloredcell{+3.55} & \coloredcell{+10.07} & \coloredcell{+2.45} & \coloredcell{+14.14} & \coloredcell{+12.93} & 49.15\% & 51.19\% & 54.25\% \\
                & Gemini-2.0-Flash  & \coloredcell{+115.91} & \coloredcell{+133.98} & \coloredcell{+166.59} & \coloredcell{+149.50} & \coloredcell{+172.02} & \coloredcell{+159.69} & \coloredcell{+61.16} & \coloredcell{+42.51} & \coloredcell{+22.84} & \coloredcell{+60.47} & \coloredcell{+55.74} & \coloredcell{+60.07} & \coloredcell{+152.66} & \coloredcell{+179.29} & \coloredcell{+232.01} & \coloredcell{+31.52} & \coloredcell{+36.44} & \coloredcell{+42.86} & 22.67\% & 22.69\% & 22.75\% \\
                & Llama-3.3-70B  & \coloredcell{-34.30} & \coloredcell{-21.37} & \coloredcell{-9.18} & \coloredcell{-48.69} & \coloredcell{-31.55} & \coloredcell{-36.65} & \coloredcell{-11.49} & \coloredcell{-12.60} & \coloredcell{-24.66} & \coloredcell{-9.23} & \coloredcell{-7.48} & \coloredcell{+2.48} & \coloredcell{-13.49} & \coloredcell{+4.17} & \coloredcell{+25.54} & \coloredcell{-82.07} & \coloredcell{-76.37} & \coloredcell{-74.83} & 9.03\% & 10.53\% & 10.60\% \\
                & Qwen-2.5-72B  & \coloredcell{+61.58} & \coloredcell{+82.01} & \coloredcell{+120.94} & \coloredcell{+68.21} & \coloredcell{+90.81} & \coloredcell{+83.77} & \coloredcell{+48.20} & \coloredcell{+27.83} & \coloredcell{+7.74} & \coloredcell{+61.26} & \coloredcell{+48.26} & \coloredcell{+59.90} & \coloredcell{+93.73} & \coloredcell{+125.27} & \coloredcell{+182.01} & \coloredcell{-12.23} & \coloredcell{-11.15} & \coloredcell{+5.44} & 19.73\% & 18.32\% & 19.77\% \\
                & Real  & 5.25 & 4.80 & 4.21 & 2.15 & 1.85 & 1.89 & 0.54 & 0.59 & 0.65 & 1.06 & 1.14 & 1.09 & 845 & 1294 & 278 & 368 & 601 & 147 & 43.55\% & 46.45\% & 52.88\% \\
            \midrule
            \multirow{6}{*}{\shortstack{Summary\\(NeurIPS 2024)}}
                & GPT-4o  & \coloredcell{+8.53} & \coloredcell{+14.44} & \coloredcell{+20.25} & \coloredcell{+38.33} & \coloredcell{+29.10} & \coloredcell{+65.36} & \coloredcell{+36.05} & \coloredcell{+18.73} & \coloredcell{+37.40} & \coloredcell{+19.46} & \coloredcell{+15.18} & \coloredcell{+23.33} & \coloredcell{+21.93} & \coloredcell{+27.41} & \coloredcell{+40.13} & \coloredcell{-26.28} & \coloredcell{-19.29} & \coloredcell{-24.44} & 23.27\% & 24.36\% & 23.99\% \\
                & Claude-3.5-Sonnet  & \coloredcell{+13.87} & \coloredcell{+10.51} & \coloredcell{+14.93} & \coloredcell{+50.66} & \coloredcell{+32.23} & \coloredcell{+64.06} & \coloredcell{+44.58} & \coloredcell{+25.75} & \coloredcell{+41.35} & \coloredcell{+14.72} & \coloredcell{+15.33} & \coloredcell{+20.53} & \coloredcell{+35.17} & \coloredcell{+24.47} & \coloredcell{+33.95} & \coloredcell{-41.45} & \coloredcell{-25.79} & \coloredcell{-27.82} & 16.68\% & 22.93\% & 23.97\% \\
                & Gemini-2.0-Flash  & \coloredcell{-6.96} & \coloredcell{-1.22} & \coloredcell{+1.97} & \coloredcell{+17.18} & \coloredcell{+9.45} & \coloredcell{+36.20} & \coloredcell{+32.72} & \coloredcell{+14.62} & \coloredcell{+25.64} & \coloredcell{+7.92} & \coloredcell{+8.68} & \coloredcell{+13.55} & \coloredcell{+12.02} & \coloredcell{+17.98} & \coloredcell{+26.09} & \coloredcell{-56.26} & \coloredcell{-51.15} & \coloredcell{-52.26} & 15.03\% & 15.92\% & 16.84\% \\
                & Llama-3.3-70B  & \coloredcell{-23.53} & \coloredcell{-13.83} & \coloredcell{-16.32} & \coloredcell{-8.81} & \coloredcell{-7.02} & \coloredcell{+2.86} & \coloredcell{+25.52} & \coloredcell{+13.20} & \coloredcell{+23.79} & \coloredcell{+0.88} & \coloredcell{+4.78} & \coloredcell{+5.96} & \coloredcell{-1.63} & \coloredcell{+10.44} & \coloredcell{+12.21} & \coloredcell{-80.42} & \coloredcell{-76.94} & \coloredcell{-80.45} & 7.66\% & 8.03\% & 7.75\% \\
                & Qwen-2.5-72B  & \coloredcell{+1.13} & \coloredcell{+3.70} & \coloredcell{+5.79} & \coloredcell{+38.00} & \coloredcell{+25.21} & \coloredcell{+39.84} & \coloredcell{+43.48} & \coloredcell{+26.56} & \coloredcell{+28.92} & \coloredcell{+14.69} & \coloredcell{+12.20} & \coloredcell{+18.82} & \coloredcell{+24.51} & \coloredcell{+23.98} & \coloredcell{+28.26} & \coloredcell{-59.61} & \coloredcell{-49.06} & \coloredcell{-44.74} & 12.49\% & 15.80\% & 19.17\% \\
                & Real  & 8.83 & 8.70 & 8.55 & 3.93 & 4.29 & 3.80 & 0.78 & 0.89 & 0.85 & 1.51 & 1.60 & 1.53 & 1473 & 2481 & 598 & 567 & 954 & 266 & 38.49\% & 38.4\%5 & 44.48\% \\
            \bottomrule
            \end{tabular}
    }

\vspace{0.3em}
\footnotesize
\textbf{Note:} Cell colors indicate the relative ratio of LLMs' performance compared to real human reviews. \textcolor{customIncreaseHigh}{Orange} shades represent higher values (over-generation), while \textcolor{customDecreaseHigh}{Green} shades represent lower values (under-generation). The intensity of the color reflects the magnitude of the deviation: light ($\pm$20--50\%), medium ($\pm$50--75\%), and dark ($>$$\pm$75\%). Uncolored cells denote deviations within $\pm$20\%.

\end{table*}

These findings reinforce the conclusion that current LLMs struggle to distinguish between papers of varying merit, both in scoring and in the structural richness of their review content. The observed differences are further illustrated in Figure~\ref{fig:num-of-node}, which shows how the number of extracted nodes varies by model and paper quality. The figure highlights the sharp structural gradients in human-written reviews and the relative flatness in LLM-generated outputs.

\subsection{Semantic Similarity Analysis}
Following the analysis of rating distributions and structural variation, we turned to a deeper question: to what extent does reviewer feedback align with the content of the original papers? To investigate this, we measured the semantic similarity between review components and paper sections. The results revealed consistent patterns across ICLR and NeurIPS, with different review components exhibiting varying degrees of alignment and clear behavioral differences between human and LLM-generated reviews.

\begin{figure*}[ht]
    \centering
    \includegraphics[width=\linewidth]{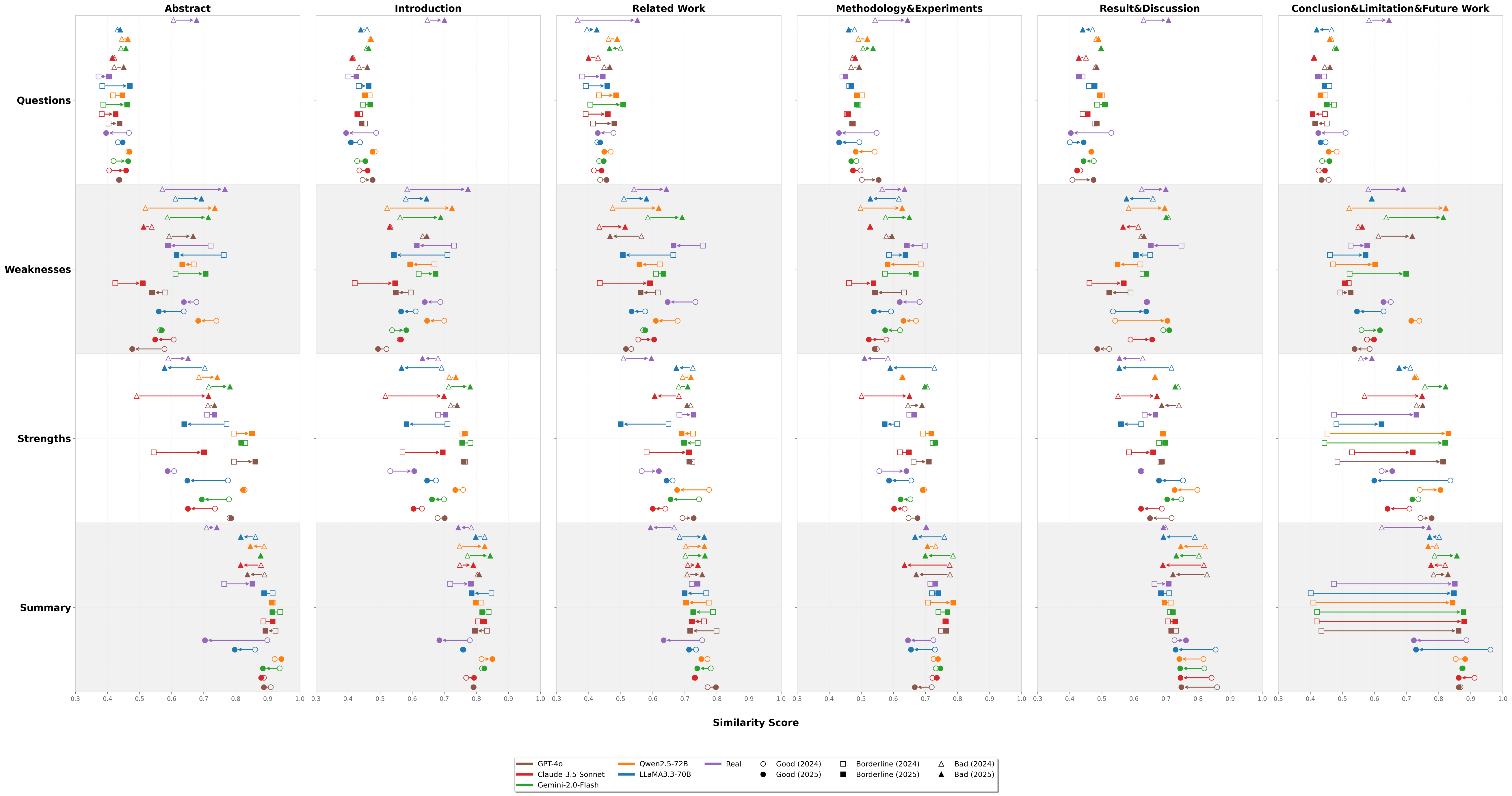}
    \caption{Semantic similarity of review components and paper sections on ICLR papers}
    \label{fig:iclr-similarity}
\end{figure*}

\begin{figure*}[!t]
    \centering
    \includegraphics[width=\linewidth]{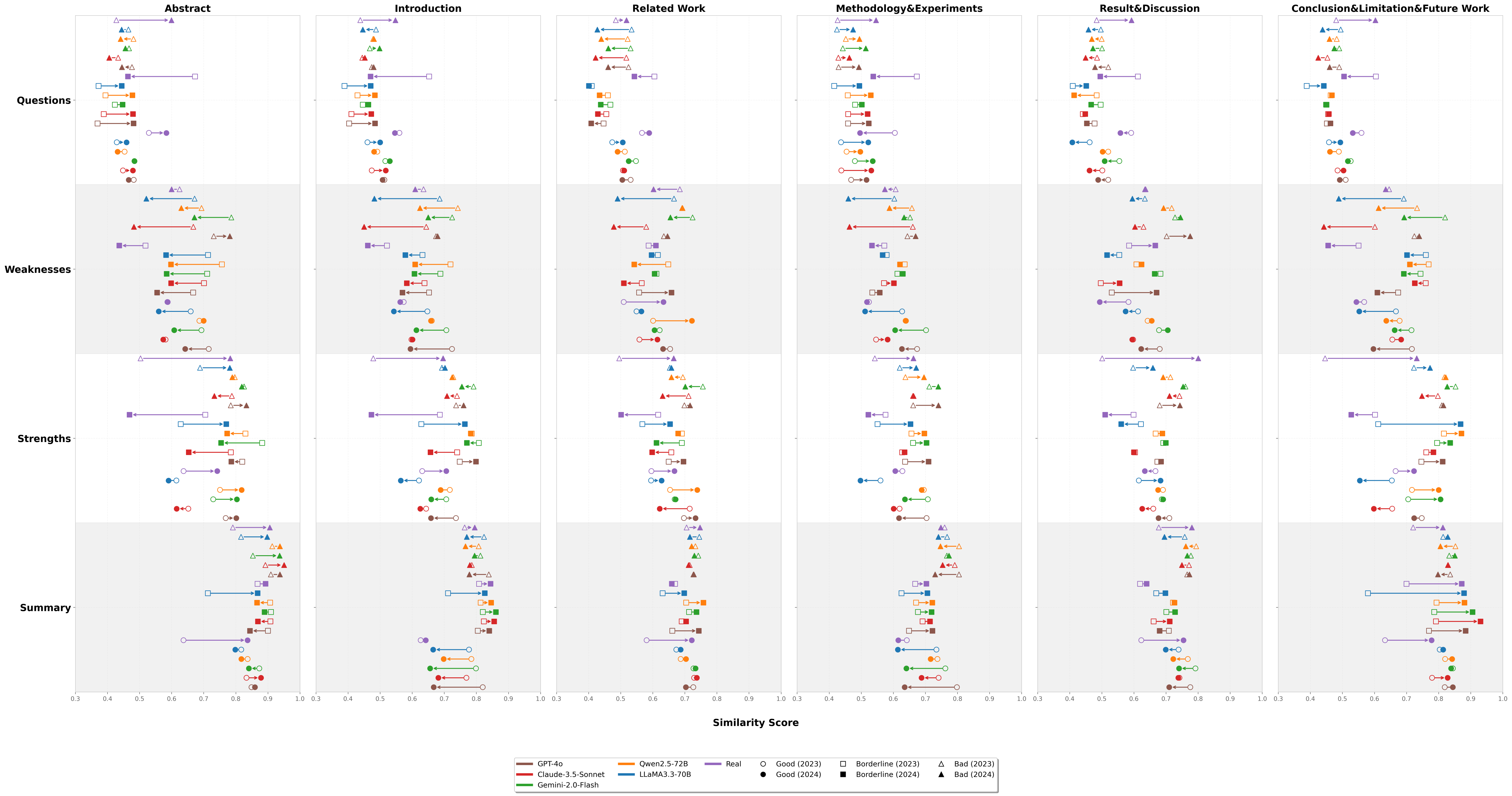}
    \caption{Semantic similarity of review components and paper sections on NeurIPS papers}
    \label{fig:neurips-similarity}
\end{figure*}

\begin{itemize}
    \item \textbf{Functional difference among review components.} We observed consistent patterns across review components in both ICLR and NeurIPS, as shown in Figure~\ref{fig:iclr-similarity} and Figure~\ref{fig:neurips-similarity}. Specifically, \textit{Summary} and \textit{Strengths} tend to exhibit relatively higher similarity scores, while \textit{Weaknesses} and \textit{Questions} show lower alignment. This pattern reflects the functional intent of each component. \textit{Summary} and \textit{Strengths} typically involve descriptive or affirmational statements grounded in explicit paper content, such as stated contributions, main results, and methodological highlights, thereby naturally producing higher similarity scores. In contrast, \textit{Weaknesses} and \textit{Questions} inherently require critical evaluation, synthesis of unstated implications, or identification of gaps, thereby necessitating the introduction of external knowledge or evaluative perspectives that diverge from the original textual content, resulting in lower semantic similarity.

    This interpretation is further supported by the \textit{In-to-Out Ratio}, defined as the number of out-of-context entities (absent in the original paper) divided by in-context entities (present in the paper). Higher ratio reflects a greater reliance on externally introduced or implicitly inferred information. As shown in Table~\ref{ICLR and NIPS Table}, both human and LLM reviews exhibit higher in-to-out ratios for \textit{Questions} and \textit{Weaknesses} than for \textit{Summary} and \textit{Strengths}. For instance, in ICLR 2025, real reviews show ratios of 62.79\% and 69.47\% for \textit{Questions} and \textit{Weaknesses}, compared to 40.62\% and 45.27\% for \textit{Summary} and \textit{Strengths}. LLMs follow the same trend; for instance, GPT-4o mirrors this trend, with even more extreme values such as 69.78\% and 98.05\% for \textit{Questions} and \textit{Weaknesses}, and only 24.09\% and 53.27\% for \textit{Summary} and \textit{Strengths}.

    \item \textbf{Behavioral divergence between LLMs and humans.} For \textit{Summary} and \textit{Strengths}, LLM-generated content generally achieves higher semantic similarity than human-written reviews. This reflects the models' tendency to closely paraphrase or replicate paper expressions, ensuring surface-level fidelity and coherence. Human reviewers, by contrast, often reframe content with broader context or interpretation, increasing semantic divergence even in descriptive sections.

    The \textit{In-to-Out Ratio} again reinforces this behavioral distinction. In the \textit{Summary} section, human-written reviews yield substantially higher ratios: 48.04\% and 43.72\% in ICLR 2024 and 2025, and 40.62\% and 38.49\% in NeurIPS 2023 and 2024, respectively. In contrast, GPT-4o's corresponding values are 26.74\%, 23.94\%, 24.09\%, and 23.27\%. A similar, though less pronounced, pattern holds for the \textit{Strengths} component. These differences indicate that human-written summaries and strengths are more likely to introduce inferred or external content beyond the literal text, while LLMs maintain closer surface-level alignment with the source material, contributing to their relatively higher semantic similarity scores.

\end{itemize}

In summary, our semantic similarity analysis highlights two key patterns: review components such as \textit{Summary} and \textit{Strengths} exhibit higher alignment with the paper due to their descriptive nature, while \textit{Weaknesses} and \textit{Questions}, which require evaluative reasoning and abstraction, show lower similarity. In addition, compared to human reviewers, LLMs are more likely to reproduce the original content with minimal abstraction, leading to higher similarity scores in descriptive sections but reduced depth in critical components.

\subsection{Knowledge Graph Analysis}
\begin{figure}[ht]
    \centering
    \includegraphics[width=0.95\linewidth]{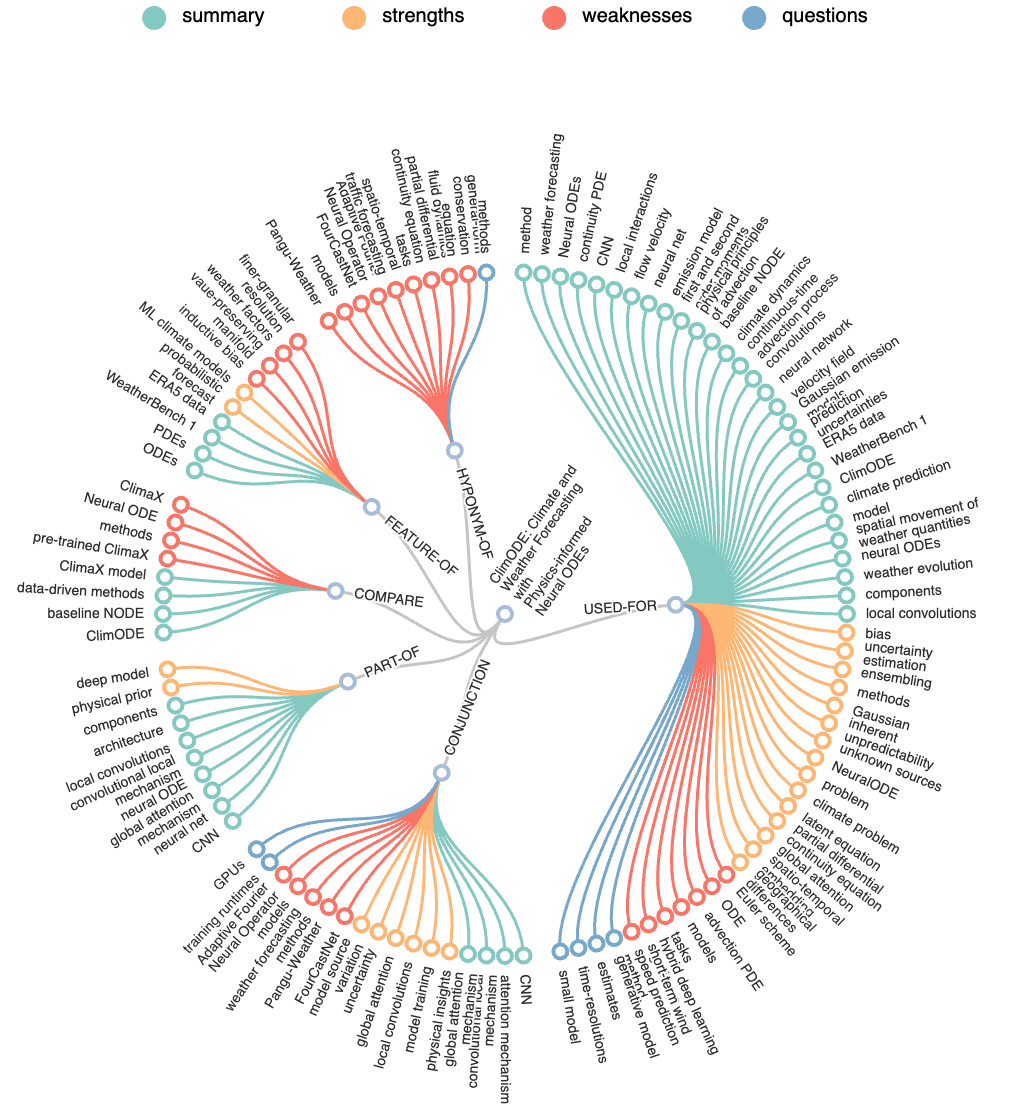}
    \caption{An example of a structured knowledge graph extracted from review components. Nodes represent scientific entities while edges encode their semantic relations. Different colors indicate review sections: \textcolor{teal}{summary}, \textcolor{orange}{strengths}, \textcolor{red}{weaknesses}, and \textcolor{blue}{questions}.}
    \label{fig:kg-graph}
\end{figure}

While semantic similarity captures surface-level alignment between reviews and paper content, it does not fully reflect the conceptual structure underlying the reviews. To address this, we constructed KGs for each review component and evaluated their structural properties to assess the depth and organization of scientific concepts expressed by both human and LLM-generated reviews. Results for ICLR (2024, 2025) and NeurIPS (2023, 2024) are presented in Tables~\ref{ICLR and NIPS Table}. Figure~\ref{fig:kg-graph} shows an example of the structured KG we constructed from review text.

\begin{itemize}
    \item \textbf{Knowledge graph structure in \textit{Weaknesses}.}
    In the \textit{Weaknesses} section, human-written reviews consistently produce richer and more conceptually dense KGs than those generated by LLMs. This is reflected in higher values across multiple structural dimensions, including the number of nodes, the number of edges, and label entropy. For example, Table~\ref{ICLR and NIPS Table} shows that among the weak papers in the ICLR 2025 dataset, real reviews contain an average of 13.68 nodes and 4.57 edges per \textit{Weaknesses} entry, whereas for LLMs, such as GPT-4o, exhibits reductions of 71.45\% in node count and 73.91\% in edge count relative to the real reviews. Similar patterns are consistently observed across other conferences and years. While some LLMs occasionally exceed real reviews in specific weaknesses or for other paper quality metrics, the overall trend remains clear: human-written \textit{Weaknesses} sections tend to integrate a greater number of scientific entities and more diverse conceptual relations. Moreover, human-written graphs in \textit{Weaknesses} show higher label entropy and greater proportions of both in-context and out-of-context entities, indicating that reviewers engage more deeply with both paper-grounded and externally inferred knowledge when identifying shortcomings.
    
    \item \textbf{Knowledge graph structure in \textit{Strengths}.}
    Conversely, the \textit{Strengths} section reveals the opposite pattern. Across most models and datasets, LLM-generated reviews tend to contain more entities and relations than their human-written counterparts. As also shown in Table~\ref{ICLR and NIPS Table}, real reviews for good papers contain an average of 6.04 nodes and 2.32 edges, while GPT-4o generates 15.74\% more nodes and 25.47\% more edges. All other LLM models except LLaMA3.3-70B also produce higher values than the real reviews. 
\end{itemize}

These results highlight a structural divergence between LLMs and human reviewers. While LLMs are more prolific in generating content in affirmational sections like \textit{Strengths}, they fall short in the conceptual depth and contextual grounding required for critical components like \textit{Weaknesses}. Human reviewers demonstrate greater knowledge diversity and abstraction when critiquing papers, highlighting their advantage in tasks demanding nuanced evaluative reasoning.

For research workflows, LLMs can serve as effective assistants in generating initial descriptive assessments, particularly in areas requiring broad coverage and recall of scientific content. Meanwhile, human reviewers should remain central for tasks demanding deeper conceptual engagement, such as detecting subtle flaws, reasoning about experimental design, or assessing broader impact. The KG structures constructed from reviews provide a complementary perspective on the organization and depth of reviewer feedback. Metrics such as the number of scientific entities, label entropy, and the ratio of in-context to out-of-context concepts capture how reviewers articulate their understanding of the paper. This structural view can assist area chairs in monitoring the quality of reviewer feedback during the decision process. Reviews that contain limited concept types, low graph complexity, or minimal critical engagement may be identified through these indicators, providing additional support in assessing whether a review sufficiently addresses the paper’s technical content and argumentative structure. By integrating these metrics into review interfaces or post-hoc diagnostic tools, conference organizers may enhance oversight and facilitate more balanced and informative evaluations across submissions.

\section{Conclusion}
This study presents a systematic evaluation of five large language models across 1,683 papers and 6,495 reviews from ICLR and NeurIPS, focusing on their alignment with human peer review. Through semantic similarity and knowledge graph analyses, we find that while LLMs excel in reproducing descriptive and affirmational content, they consistently underperform in critical dimensions such as identifying weaknesses, raising substantive questions, and modulating feedback based on submission quality. The high-consensus benchmark and structured evaluation framework introduced in this work offer not only practical resources but also new methodological perspectives for assessing the quality of LLM-generated reviews, paving the way for future research on more discerning and context-aware automated reviewers.

This study also has limitations. First, the benchmark dataset is limited to papers from ICLR and NeurIPS, which may constrain the generalizability of our findings to other venues with different disciplinary focuses, review practices, or evaluation standards. Second, we only investigate five LLMs and adopt a fixed prompt design based on official rubrics, which limits the representativeness of our findings across rapidly evolving models and restricts insights of how prompt variations influence model performances, especially in evaluative tasks requiring nuanced critique or quality-sensitive reasoning. Additionally, scientific disagreements are inherent in the peer review process. To evaluate the capabilities of LLMs, our dataset only includes papers where reviewers reached high consensus, excluding those with disagreements due to the lack of reliable evaluation standards. Future work could address these limitations by incorporating diverse conferences, journals, and interdisciplinary domains to enhance generalizability. Further exploration of newer or domain-specific models, alternative prompting strategies, and fine-tuning methods may also yield a deeper understanding of LLM performance in generating high-quality reviews.

\bibliographystyle{ieeetr}
\bibliography{mybibliography}

\end{document}